\pdfoutput=1
\documentclass[11pt]{article}
\usepackage[preprint]{acl}
\usepackage{times}
\usepackage{latexsym}
\usepackage[T1]{fontenc}
\usepackage[utf8]{inputenc}
\usepackage{amsmath}
\usepackage{amssymb}  
\usepackage{microtype}
\usepackage{inconsolata}
\usepackage{graphicx}
\usepackage{booktabs}
\usepackage{multirow}
\usepackage{tcolorbox}
\usepackage{hyperref}
\usepackage[table,xcdraw]{xcolor}
\usepackage[normalem]{ulem}
\useunder{\uline}{\ul}{}
\title{Controlling and Assessing Appropriate Persona Use\\in LLM-based Dialogue Generation}

\author{
  Jongkyung Shin\textsuperscript{1}\footnotemark[1]\footnotemark[2]\;
  Inkyu Lee\textsuperscript{2}\footnotemark[1]\;
  Chiehyeon Lim\textsuperscript{1,2,3}\footnotemark[2]\;
  \\
  \\
    \textsuperscript{1}UNIST\quad
    \textsuperscript{2}POSTECH\quad
    \textsuperscript{3}POSCO Holdings Inc.\quad
  \\
  \small\texttt{shinjk1156@unist.ac.kr, \{wer070947,chiehyeon.lim\}@postech.ac.kr}
}

\begin{document}
\maketitle
{\let\thefootnote\relax\footnotetext{$*$: Co-first authors, $\dagger$: Co-corresponding authors.}}
\begin{abstract}
In persona-based dialogue generation (PDG), LLMs often overuse persona attributes by incorporating them regardless of dialogue context, resulting in unnatural responses. Despite its practical significance, the underlying causes remain unexplored, with no method to mitigate this problem or metric to assess the appropriateness of persona use. To address these issues, we first conduct a comprehensive analysis of LLM-based PDG, revealing that LLMs exhibit a systematic bias to incorporate all given persona attributes, and that existing metrics fail to capture contextual appropriateness. Building on these findings, we propose Self-CONtrastive Persona Overuse Suppression (SCONPOS) to mitigate overuse by directly intervening in LLMs' internal representations at the prompt encoding stage, without requiring any response generation. We further propose the Persona Appropriateness Score (PAS), a novel metric that penalizes both overuse and underuse. Experimental results demonstrate that SCONPOS systematically reduces overuse, and PAS captures the contextual appropriateness of persona use.
\end{abstract}

\section{Introduction}
Persona-based dialogue generation (PDG) aims to generate responses that are not only contextually appropriate and fluent, but also faithful to a predefined persona. By enabling conversational agents to embody particular roles, traits, or identities, PDG supports the development of systems tailored to diverse interaction goals and user needs. As a result, PDG has shown promise in a wide range of applications, including mental health counseling \cite{de2020effectiveness, hong2024expanding}, game-based interaction \cite{weir-etal-2024-ontologically, nananukul2024if}, and educational assistance \cite{liu-etal-2024-personality, rooein-etal-2026-pats}.

\begin{figure}[t]
    \centering
    \includegraphics[width=\columnwidth]{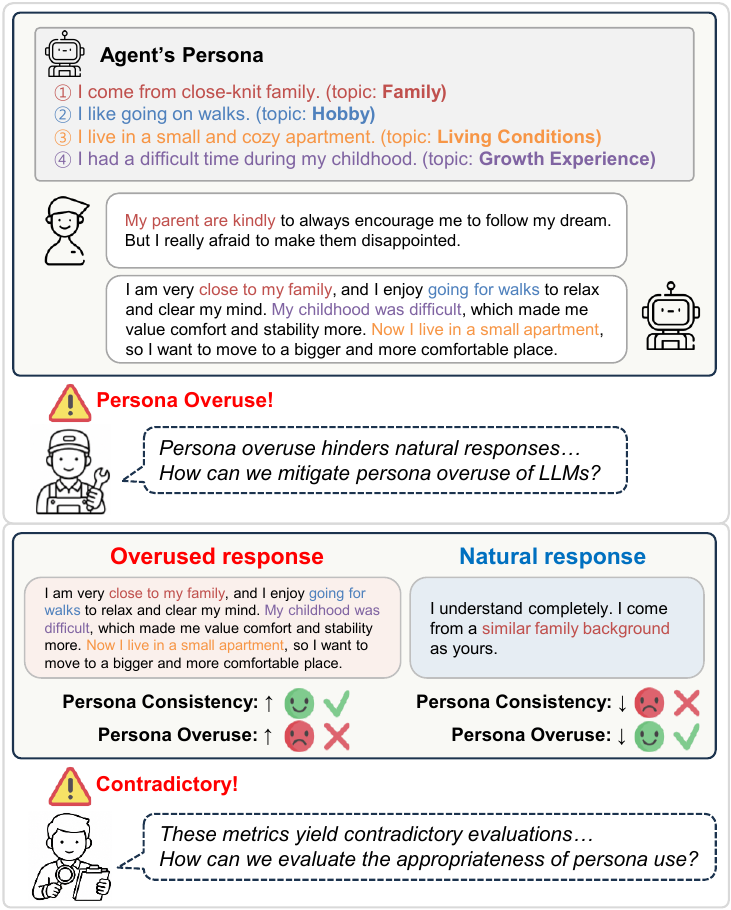}
    \caption{The motivation of our work, highlighting the critical need for: (Top) mitigating the persona overuse problem in LLM-based PDG, and (Bottom) developing a metric to assess the appropriateness of persona use.}
    \label{fig:1}
\end{figure}

Early studies on PDG adopted learning-based approaches with moderately sized pre-trained language models \cite{li-etal-2020-dont, song-etal-2021-bob}. However, these methods often exhibited limited persona consistency, in part because available persona dialogue datasets lacked sufficient scale and diversity \cite{hong-etal-2025-dialogue, jandaghi-etal-2024-faithful}. With recent advances in large language models (LLMs), their enhanced instruction following capabilities have enabled high-quality PDG even without explicit parameter updates \cite{NEURIPS2020_1457c0d6, 10.1145/3560815}. In particular, in-context learning (ICL)-based approaches have demonstrated strong persona consistency and generalization to unseen personas \cite{xu-etal-2023-towards-zero, ni-etal-2023-multi, pu2024crafting}. This allows dialogue agents to be cost-effectively adapted to new personas by simply modifying the prompt.

However, recent studies have reported a persona overuse problem, where LLMs overemphasize the assigned persona regardless of the contextual relevance \cite{kim-etal-2024-panda}. This leads to unnatural responses that disrupt dialogue flow. In real-world applications such as mental health counseling and customer service, where careful and context-sensitive responses are essential, such behavior can severely undermine the reliability of the interaction and user trust of the system. Despite its practical importance, the mechanisms for mitigating persona overuse in LLM-based PDG remain underexplored.

Meanwhile, attempts to mitigate persona overuse can introduce persona underuse, where responses fail to reflect persona attributes, thereby reducing persona consistency. Although these are opposite phenomena, they stem from a shared underlying issue: persona attributes are not expressed appropriately given the dialogue context. From this perspective, the central challenge of PDG is not merely to reflect persona attributes as much as possible, but to control when and to what extent they should be reflected. However, there is still no established metric for quantitatively assessing the appropriateness of persona use, highlighting a critical limitation in current PDG evaluation. 

To fill these research gaps, our work makes the following contributions:
\paragraph{Comprehensive Analysis.} We conduct a systematic analysis of persona use in LLM-based PDG. Our analysis reveals that LLMs exhibit an overuse-inducing bias to incorporate all given persona attributes regardless of dialogue context, and the existing metrics fail to capture contextual appropriateness of persona use.

\paragraph{Overuse Mitigation Method.} We propose Self-CONtrastive Persona Overuse Suppression (SCONPOS), the first method to mitigate persona overuse by suppressing overuse-inducing bias in LLMs' internal representations without requiring response generation. Experimental results demonstrate that SCONPOS effectively and efficiently reduces persona overuse across datasets and models.

\paragraph{Appropriateness Evaluation Metric.} We propose Persona Appropriateness Score (PAS), the first metric for assessing the appropriateness of persona use in PDG. Our experiments demonstrate that PAS captures the degree to which persona attributes are contextually appropriate.

\section{Analysis}\label{sec:2}
We first conduct a systematic analysis to address the following questions: (Q1) Do LLMs have the ability to control persona use depending on the partner's utterance? (Q2) Can existing metrics assess the appropriateness of persona use?

\begin{table*}[]
\centering
\setlength{\tabcolsep}{3pt}
\resizebox{\textwidth}{!}{%
\begin{tabular}{@{}cclccccccccccccccccccc@{}}
\toprule
\multirow{3}{*}{\begin{tabular}[c]{@{}c@{}}Utterance\\ type\end{tabular}} & \multirow{3}{*}{\begin{tabular}[c]{@{}c@{}}Persona\\ setting\end{tabular}} & \multicolumn{10}{c}{PersonaChat} &  & \multicolumn{9}{c}{MBTI-S2Conv} \\ \cmidrule(lr){4-12} \cmidrule(l){14-22} 
 &  &  & \multirow{2}{*}{\textbf{Overuse} ($\downarrow$)} &  & \multicolumn{3}{c}{Consistency ($\uparrow$)} &  & \multicolumn{3}{c}{Ref. Similarity ($\uparrow$)} &  & \multirow{2}{*}{\textbf{Overuse} ($\downarrow$)} &  & \multicolumn{3}{c}{Consistency ($\uparrow$)} & \textbf{} & \multicolumn{3}{c}{Ref. Similarity ($\uparrow$)} \\ \cmidrule(lr){6-8} \cmidrule(lr){10-12} \cmidrule(lr){16-18} \cmidrule(l){20-22} 
 &  &  &  &  & C.Score & P-Dist & P-F1 &  & BLEU & R-L & B.Score &  &  &  &  & P-Dist & P-F1 &  & BLEU & R-L & B.Score \\ \midrule
\multirow{3}{*}{\begin{tabular}[c]{@{}c@{}}Persona\\ irrelevant\end{tabular}} & \textbf{No} &  & \textbf{0.598} &  & 0.112 & 0.333 & 0.052 &  & \textbf{0.043} & \textbf{0.097} & \textbf{0.833} &  & \textbf{0.344} &  &  & 0.235 & 0.084 &  & \textbf{0.135} & \textbf{0.163} & \textbf{0.851} \\
 & \textbf{+Noise} &  & 0.822 &  & 0.517 & 0.399 & 0.067 &  & \textbf{0.043} & 0.096 & \textbf{0.833} &  & 0.527 &  &  & 0.263 & 0.093 &  & 0.122 & 0.146 & 0.849 \\
 & \textbf{All} &  & 0.898 &  & \textbf{0.919} & \textbf{0.496} & \textbf{0.074} &  & 0.042 & 0.091 & 0.831 &  & 0.616 &  &  & \textbf{0.273} & \textbf{0.096} &  & 0.116 & 0.146 & 0.848 \\ \midrule
\multirow{4}{*}{\begin{tabular}[c]{@{}c@{}}Persona\\ relevant\end{tabular}} & \textbf{No} &  & \textbf{0.666} &  & 0.070 & 0.356 & 0.053 &  & 0.044 & 0.090 & 0.834 &  & \textbf{0.858} &  &  & 0.347 & 0.115 &  & \textbf{0.108} & \textbf{0.122} & \textbf{0.854} \\
 & \textbf{Match} &  & 0.753 &  & 0.622 & 0.428 & 0.070 &  & \textbf{0.046} & \textbf{0.095} & \textbf{0.835} &  & 0.866 &  &  & 0.359 & \textbf{0.116} &  & 0.105 & 0.120 & 0.852 \\
 & \textbf{+Noise} &  & 0.817 &  & 0.754 & 0.463 & 0.073 &  & 0.045 & 0.094 & 0.834 &  & 0.878 &  &  & 0.362 & \textbf{0.116} &  & 0.103 & 0.119 & 0.852 \\
 & \textbf{All} &  & 0.859 &  & \textbf{0.868} & \textbf{0.499} & \textbf{0.075} &  & 0.045 & 0.093 & 0.833 &  & 0.890 &  &  & \textbf{0.366} & 0.115 &  & 0.101 & 0.117 & 0.851 \\ \bottomrule
\end{tabular}%
}
\caption{Results of the analysis experiments. Scores are averaged across Llama3.1, Qwen2.5, Deepseek-chat, and GPT4.1, while the results for each model are reported in Tables \ref{append_table:irrelevant_analysis_agent_only_setting} and \ref{append_table:relevant_analysis_agent_only_setting} in the Appendix. The best value for each metric is shown in bold. ‘R-L’ and ‘B.Score’ indicate ROUGE-L and BERTScore, respectively.}
\label{table:1}
\vspace{-5pt}
\end{table*}

\subsection{Analysis setup}
If LLMs can appropriately control persona use, then even when persona attributes irrelevant to the current conversation topic are added, they should selectively reflect only those persona attributes that align with that topic. Accordingly, the persona use expressed in the response should not differ substantially across settings that vary only in the combination of attributes drawn from the same persona. To examine this, following \citet{kim-etal-2024-panda}, we compare the topic of the partner's last utterance with the topic of the persona attribute reflected in the response. We introduce four persona input settings. \textbf{All} corresponds to the original PDG setting, in which all predefined persona attributes are given to the LLM. \textbf{Match} is the setting in which only persona attributes that match the topic of the partner's last utterance are provided. \textbf{+Noise} extends the \textbf{Match} by additionally providing one persona attribute that does not align with the utterance topic. \textbf{No} is the setting in which no persona is provided.

Furthermore, because the appropriateness of persona use may vary depending on the dialogue context, we divide partner utterances into two types. \textit{Persona-relevant utterances} are those whose topic matches at least one of the predefined persona attributes. In such cases, incorporating the corresponding attribute into a response is contextually natural and appropriate. In contrast, \textit{persona-irrelevant utterances} are those whose topic does not match any persona attribute, in which case a response that does not reflect persona is more appropriate. For example, if the topic of the partner’s utterance is \textit{hobby} but the given persona contains no related attribute, it is more appropriate to continue the conversation naturally on that topic rather than incorporating an unrelated persona attribute.

We conduct experiments on two datasets: PersonaChat \cite{zhang-etal-2018-personalizing} with short and single sentence persona attributes, and MBTI-S2Conv \cite{tu2023characterchat} with more detailed and descriptive personas, across four models: Llama3.1 \cite{grattafiori2024llama}, Qwen2.5 \cite{qwen2025qwen25technicalreport}, DeepSeek-chat \cite{bi2024deepseek}, and GPT4.1 \cite{achiam2023gpt}. To analyze responses, we measure \textbf{Overuse} \cite{kim-etal-2024-panda}, which captures the use of persona attributes that are contextually irrelevant to the partner's utterance, as well as the overabundance of persona attributes used in a response. For persona consistency, we report C.Score\footnote{C.Score is omitted for MBTI-S2Conv because it requires a dataset-specific evaluator trained on the dialogue NLI dataset \cite{welleck-etal-2019-dialogue}, which is unavailable for MBTI-S2Conv.} \cite{madotto-etal-2019-personalizing}, P-Dist \cite{cho-etal-2022-personalized}, and P-F1 \cite{lian2019learning}. We further use BLEU \cite{papineni-etal-2002-bleu}, ROUGE-L \cite{lin-2004-rouge}, and BERTScore \cite{zhang2019bertscore}, which evaluate similarity to the reference. Detailed analysis settings are provided in Appendix~\ref{pas_setup}. In addition, experimental results for the setting that additionally includes the partner's persona are presented in Appendix~\ref{anal_partner}.

\subsection{Results and Implications}
As shown in Table \ref{table:1}, persona use in the responses differed considerably across persona input settings. Across both datasets, \textbf{Overuse} increased as more persona attributes unrelated to the topic of the partner's utterance were added, indicating that LLMs tend to incorporate as many given persona attributes as possible into their responses rather than selectively using contextually relevant ones. The same pattern was observed when both the agent's persona and the partner's persona were provided (See Tables \ref{append_table:irrelevant_analysis_partner_agent_setting} and \ref{append_table:relevant_analysis_partner_agent_setting} in the Appendix). This tendency may stem from the way LLMs are typically trained, where models are optimized to satisfy all provided conditions, leading them to treat persona attributes as instructions to follow rather than as information to be used selectively based on the dialogue context. This suggests that an overuse-inducing bias is already formed at the stage where the model encodes the prompt.

If a metric can assess the appropriateness of persona use, it should be able to penalize both overuse and underuse. However, existing persona consistency metrics assigned the best scores to the \textbf{All} setting, where overuse was most pronounced\footnote{P-F1 computes the proportion of persona-related token overlap in the response, making it sensitive to response length. In MBTI-S2Conv, responses to persona-relevant utterances were on average longer than in PersonaChat (\textbf{No}: 110 vs. 57 tokens; \textbf{All}: 134 vs. 70 tokens), which dilutes the token overlap ratio and suppresses P-F1 scores. This explains the exception where the \textbf{All} setting did not achieve the best P-F1 score for persona-relevant utterances in MBTI-S2Conv, despite containing the most persona information.}, and the worst scores to the \textbf{No} setting for persona-irrelevant utterances, even though responses that do not reflect persona may be more natural in that context. This indicates that these metrics reward responses that include more persona information, regardless of the utterance context. Moreover, in persona-relevant utterances, \textbf{Overuse} assigned the best score to the \textbf{No} setting, even though its responses did not include the necessary persona attributes. Reference-based metrics failed to adequately reflect differences in persona use across settings, despite the responses varying. In summary, persona consistency metrics do not appropriately penalize overuse, \textbf{Overuse} does not capture underuse, and reference-based metrics do not sufficiently distinguish differences across settings. Consequently, existing metrics are insufficient to assess the appropriateness of persona use.

\begin{figure*}
    \centering
    \includegraphics[width=\linewidth]{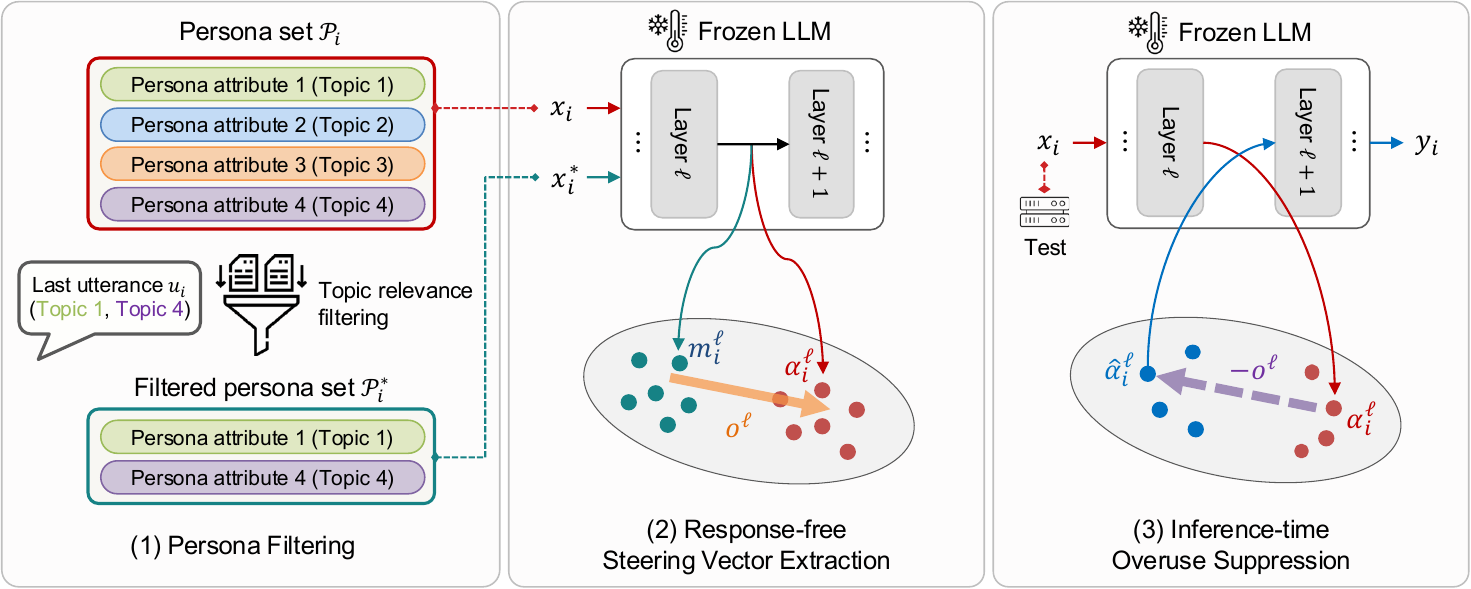}
    \caption{Overview of SCONPOS.}
    \label{fig:overview_self_conpos}
\end{figure*}

\section{SCONPOS: Self-CONtrastive Persona Overuse Suppression}
Motivated by the insights from Section \ref{sec:2}, we propose SCONPOS, the first method to mitigate persona overuse in LLM-based PDG. Our core idea is to directly suppress the overuse-inducing bias formed at the prompt encoding stage in the internal representation space. Through this intervention, we aim to enable the model to selectively reflect only the persona attributes that are appropriate for the current dialogue context. Figure \ref{fig:overview_self_conpos} shows an overview of our method.

\subsection{Persona Filtering}
The tendency of LLMs to reflect all given persona attributes can serve as a clue for identifying the internal representations associated with overuse. The difference in activations between the \textbf{All} and \textbf{Match} settings can capture the representational components induced by context-irrelevant persona conditioning, i.e., the overuse-inducing bias.

To this end, we filter the full persona set $\mathcal{P}_i$ based on the topics of the partner's last utterance $u_i$ to construct a context-appropriate persona set $\mathcal{P}_i^*$:
\begin{equation}
    \mathcal{P}_i^* = \left\{ p \in \mathcal{P}_i \mid \text{topic}(p) \in \text{topics}(u_i) \right\}.
\end{equation}
The resulting pair $(x_i, x_i^*)$, where $x_i = (\mathcal{P}_i, H_i, u_i)$ and $x_i^* = (\mathcal{P}_i^*, H_i, u_i)$, serves as a self-contrastive pair, with $H_i$ denoting the dialogue history.

\subsection{Response-free Steering Vector Extraction}
For each self-contrastive pair $(x_i, x_i^*)$, we extract the activations $\alpha_i^l$ and $m_i^l$ at the last prompt token of layer $l$ from a frozen LLM. We then compute the overuse-inducing vector $o^l$ by averaging the differences across the dataset $\mathcal{D} = \{x_i\}_{i=1}^{N}$:
\begin{equation}
    o^l = \frac{1}{N} \sum_{i=1}^{N} \left( \alpha_i^l - m_i^l \right).
\end{equation}
We emphasize that this extraction relies solely on the prompt $x$ without any response $y$. Unlike conventional representation engineering methods that require response generation for vector extraction \cite{rimsky-etal-2024-steering, zhang-etal-2025-personalized}, our approach operates in a response-free manner.

\subsection{Inference-time Overuse Suppression}
We suppress the overuse-inducing bias by subtracting $o^l$ from the hidden state $\alpha_i^l$ at layer $l$ during inference, including response generation:
\begin{equation}
    \hat{\alpha}_i^l = \alpha_i^l - \lambda o^l
\end{equation}
where $\lambda \ge 0$ is a hyperparameter to control the suppression intensity, and $\hat{\alpha}_i^l$ is passed as input to layer $l+1$. Notably, this intervention is performed only at a specific layer $l$, and requires only a single subtraction operation with no additional parameter updates or complex computations.

\section{PAS: Persona Appropriateness Score}
To address the limitations of existing metrics as discussed in Section \ref{sec:2}, we propose a novel metric, PAS. Unlike counting-based metrics such as \textbf{Overuse} and C.Score, PAS measures the discrepancy between the topic distributions of the partner's utterance and the model's response to evaluate the appropriateness of persona use in the dialogue context. The key intuition is that the more the topic distribution of the response aligns with the persona-relevant context implied by the partner's utterance, the more appropriately the persona is used.

We define the topic space $T_{ps}$ as the set of topics associated with the predefined personas, so that PAS captures persona-specific topical alignment rather than general topical coherence between the utterance and response. We additionally introduce a dummy topic $t_0$ into $T_{ps}$ to handle cases where the partner's utterance or the model's response does not correspond to any predefined persona topic. Subsequently, we extract topics from the partner's last utterance and the model's response, and represent their usage as discrete probability distributions, $P_i^u$ and $P_i^r$, within the topic space $T_{ps}$.
For each instance $i$, $\textrm{PAS}_i$ is defined as follows:
\begin{equation}
    \textrm{PAS}_i = 1 - \sqrt{ \frac{D_{\mathrm{KL}}(P_{i}^u \| M_i) + D_{\mathrm{KL}}(P_{i}^r \| M_i)}{2} }
\end{equation}
where $M_i=\frac{P_i^u + P_i^r}{2}$ and $D_{\mathrm{KL}}$ is the Kullback-Leibler divergence between two topic distributions. PAS employs the Jensen-Shannon Distance \cite{1207388} for two reasons. First, it satisfies the metric properties and admits an information-theoretic interpretation. Second, it yields a finite and stable measure even when the two distributions do not overlap, making it widely used for comparing categorical topic distributions \cite{schlechtweg-etal-2019-wind, bommasani-cardie-2020-intrinsic, markoski-etal-2021-cultural}.

The overall PAS is computed by separately averaging $\textrm{PAS}_i$ over persona-relevant and persona-irrelevant utterances, and taking their harmonic mean. This is because a model that never reflects persona attributes can still achieve high scores on persona-irrelevant utterances, inflating the overall score. Accordingly, a high score requires appropriate persona use in both cases.
\begin{equation}
    \textrm{PAS} = \frac{2 \cdot \textrm{PAS}_{\text{rel}} \cdot \textrm{PAS}_{\text{irrel}}}{\textrm{PAS}_{\text{rel}} + \textrm{PAS}_{\text{irrel}}}
\end{equation}
where $\textrm{PAS}_{\text{rel}}$ and $\textrm{PAS}_{\text{irrel}}$ are the average $\textrm{PAS}_i$ over persona-relevant and persona-irrelevant utterances, respectively. Intuitively, PAS penalizes both the case where irrelevant persona attributes are mentioned (overuse) and the case where relevant persona topics are ignored (underuse).

\section{Experiment}
\subsection{Setup}
\paragraph{Data and model.} We conduct experiments on two datasets, PersonaChat and MBTI-S2Conv. Each dataset is split into train/valid/test sets with a ratio of 0.72/0.08/0.2. We experiment with three models: Llama3.1, Qwen2.5, and DeepSeek-chat.

\paragraph{Baselines.} We compare SCONPOS against ICL-based methods including Chain-of-Thought (CoT) \cite{kojima2022large}, Task decomposition (Decomp.) \cite{khot2022decomposed}, and Self-refine \cite{madaan2023self}, which control persona use at the prompt level, and CAA \cite{rimsky-etal-2024-steering}, which applies steering vectors after the prompt tokens and requires response generation for steering vector extraction. 
\paragraph{Implementation details.} The suppression intensity $\lambda$ and target layer $l$ of SCONPOS are set to the values that maximize the harmonic mean of ROUGE-L and PAS on the validation set, balancing generation quality and persona appropriateness. The steering vector is extracted from the hidden state of the last prompt token. We adopt the vanilla prompt, as shown in Figure \ref{fig:prompt_vanilla}. For a competitive comparison, we use responses generated in the All and Match settings to extract the steering vector for CAA. Other details including human evaluation are provided in Appendix \ref{append:sec:B}. The implementation of SCONPOS and PAS is publicly available at \url{https://github.com/LIK9/SCONPOS-PAS}.

\subsection{Experimental results of SCONPOS}
\paragraph{Does SCONPOS effectively mitigate overuse?}
We conduct a performance comparison to verify the effectiveness of SCONPOS. As shown in Table \ref{tab:performance_comparision}, SCONPOS systematically reduces overuse across both short-form and descriptive long-form personas. In contrast, ICL-based methods show limited effectiveness in mitigating overuse, and CAA also reduces overuse but performs lower than SCONPOS.

\begin{table}[h]
\centering
\setlength{\tabcolsep}{3pt}
\resizebox{\columnwidth}{!}{%
\begin{tabular}{@{}lcccccc@{}}
\toprule
 & Vanilla & CoT & Decomp. & Self-refine & CAA & SCONPOS \\ \midrule
\multicolumn{7}{l}{\textit{PersonaChat}} \\ \midrule
Overuse ($\downarrow$) & 0.917 & 0.867 & 0.899 & 0.913 & {\ul0.830} & \textbf{0.743} \\
ROUGE-L ($\uparrow$) & 0.086 & 0.090 & 0.090 & 0.079 & {\ul 0.091} & \textbf{0.103} \\
PAS ($\uparrow$) & 0.049 & 0.128 & 0.081 & 0.086 & {\ul 0.224} & \textbf{0.301} \\ \midrule
\multicolumn{7}{l}{\textit{MBTI-S2Conv}} \\ \midrule
Overuse ($\downarrow$) & 0.861 & 0.819 & 0.864 & 0.848 & {\ul0.816} & \textbf{0.722} \\
ROUGE-L ($\uparrow$) & 0.120 & 0.120 & 0.116 & 0.111 & {\ul0.127} & \textbf{0.135} \\
PAS ($\uparrow$) & 0.322 & 0.402 & 0.326 & 0.335 & {\ul0.423} & \textbf{0.540} \\ \bottomrule
\end{tabular}%
}
\caption{Performance comparison results of Llama3.1. Results for other models are in Table~\ref{tab:full_performance_comparison}.}
\label{tab:performance_comparision}
\vspace{-10pt}
\end{table}

\paragraph{Does the steering vector capture overuse-inducing bias?}
To verify whether the extracted steering vector actually captures overuse-inducing bias, we conduct cross-dataset transfer experiments by applying a steering vector extracted from one dataset to another. As shown in Table \ref{tab:cross_dataset_transferability}, SCONPOS consistently reduces overuse and improves ROUGE-L and PAS across both transfer directions. In contrast, CAA fails to improve performance when the vector extracted from MBTI-S2Conv is applied to PersonaChat, and shows lower performance than SCONPOS. This implies that vectors extracted at the response level are limited to capturing surface-level patterns specific to each dataset. On the other hand, the transferability of SCONPOS suggests that it captures the overuse-inducing bias inherent in LLMs, rather than patterns specific to each dataset.

\begin{table}[!t]
\centering
\resizebox{\columnwidth}{!}{%
\begin{tabular}{@{}lccc@{}}
\toprule
 & Vanilla & CAA & SCONPOS \\
\midrule
\multicolumn{4}{l}{\textit{Extract: MBTI-S2Conv $\to$ Steer: PersonaChat}} \\
\midrule
Overuse ($\downarrow$)  & 0.917 & 0.901 (-2\%)  & \textbf{0.811 (-12\%)} \\
ROUGE-L ($\uparrow$)   & 0.086 & 0.094 (+9\%)  & \textbf{0.111 (+28\%)} \\
PAS ($\uparrow$)        & 0.049 & 0.049 (0\%)   & \textbf{0.206 (+320\%)} \\
\midrule
\multicolumn{4}{l}{\textit{Extract: PersonaChat $\to$ Steer: MBTI-S2Conv}} \\
\midrule
Overuse ($\downarrow$)  & 0.861 & 0.825 (-4\%)  & \textbf{0.791 (-8\%)} \\
ROUGE-L ($\uparrow$)   & 0.120 & 0.117 (-3\%)  & \textbf{0.135 (+12\%)} \\
PAS ($\uparrow$)        & 0.322 & 0.385 (+19\%) & \textbf{0.455 (+41\%)} \\
\bottomrule
\end{tabular}%
}
\caption{Cross-dataset transferability of CAA and SCONPOS with Llama3.1. Percentages indicate relative change from the Vanilla prompt.  Results for other models are in Table~\ref{tab:cross_dataset}.
}
\vspace{-10pt}
\label{tab:cross_dataset_transferability}
\end{table}

\begin{table}[b]
\centering
\resizebox{\columnwidth}{!}{%
\begin{tabular}{@{}lccc@{}}
\toprule
 & Vanilla (All) & SCONPOS & SCONPOS (dup.) \\
\midrule
\multicolumn{4}{l}{\textit{PersonaChat}} \\
\midrule
Overuse ($\downarrow$) & 0.821 & 0.649 & 0.628 \\
ROUGE-L ($\uparrow$) & 0.091 & 0.105 & 0.103 \\
PAS ($\uparrow$)       & 0.422 & 0.441 & 0.419 \\
\midrule
\multicolumn{4}{l}{\textit{MBTI-S2Conv}} \\
\midrule
Overuse ($\downarrow$) & 0.882 & 0.817 & 0.813 \\
ROUGE-L ($\uparrow$) & 0.123 & 0.135 & 0.141 \\
PAS ($\uparrow$)       & 0.362 & 0.412 & 0.399 \\
\bottomrule
\end{tabular}%
}
\caption{Performance with controlled attribute count and prompt length. (dup.) denotes duplicated relevant attributes in the Match setting. Evaluated on persona-relevant utterances, averaged across three models.}
\label{tab:control_vector}
\end{table}

Additionally, since the Match setting contains fewer attributes than the All setting, the steering effect could stem from differences in attribute count or prompt length. To control for this, we duplicate the relevant attributes in the Match setting so that both properties approximate those of the All setting. This duplication is applicable only when relevant attributes exist, so we evaluate on persona-relevant utterances. As shown in Table~\ref{tab:control_vector}, the controlled vector achieves comparable overuse reduction to SCONPOS, indicating that the steering effect is not attributable to surface-level prompt properties.

\paragraph{Is SCONPOS more efficient than CAA?}
To verify the practical advantages of SCONPOS, we compare its efficiency with CAA. As shown in Table \ref{tab:cost_efficiency}, SCONPOS is approximately 4.5$\times$ faster than CAA in total extraction time. This difference arises from two points. First, CAA requires response generation for All and Match settings prior to vector extraction. Second, the vector extraction process requires both the prompt and the generated response as input, incurring additional time. In contrast, SCONPOS extracts steering vectors from the prompt alone without response generation, incurring neither of these additional costs. Meanwhile, inference time is comparable for both methods. SCONPOS requires only 20KB of storage to suppress overuse-inducing bias. This demonstrates that SCONPOS achieves practical overuse mitigation in a time- and cost-efficient manner.

\begin{table}[h]
\centering
\resizebox{0.95\columnwidth}{!}{%
\begin{tabular}{@{}llcc@{}}
\toprule
\multicolumn{2}{c}{Cost} & CAA & SCONPOS \\ \midrule
\multicolumn{2}{l}{Total extraction time (min)} & 254.2 & 56.9 \\
 & Response generation time & 180.5 & - \\
 & Vector extraction time & 73.7 & 56.9 \\
\multicolumn{2}{l}{Inference time (sec/it)} & 4.32 & 4.20 \\
\multicolumn{2}{l}{Storage space (KB)} & 20 & 20 \\ \bottomrule
\end{tabular}%
}
\caption{Time and storage cost comparison between CAA and SCONPOS with MBTI-S2Conv dataset. Per-model results are in Table \ref{tab:cost_efficiency_full}.}
\vspace{-10pt}
\label{tab:cost_efficiency}
\end{table}

\begin{table}[!b]
\centering
\resizebox{0.99\columnwidth}{!}{%
\begin{tabular}{@{}cccc@{}}
\toprule
Dataset & ICL Method & \begin{tabular}[c]{@{}c@{}}Persona \\ Appropriateness\end{tabular} & \begin{tabular}[c]{@{}c@{}}Dialogue \\ Coherence\end{tabular} \\ \midrule
\multirow{4}{*}{PersonaChat} & Vanilla & 0.76 & 0.68 \\
 & CoT & 0.70 & 0.60 \\
 & Decomp. & 0.74 & 0.72 \\
 & Self-refine & 0.66 & 0.58 \\ \midrule
\multirow{4}{*}{MBTI-S2Conv} & Vanilla & 0.68 & 0.60 \\
 & CoT & 0.62 & 0.64 \\
 & Decomp. & 0.60 & 0.66 \\
 & Self-refine & 0.64 & 0.62 \\ \bottomrule
\end{tabular}%
}
\caption{Win rates of ICL methods with SCONPOS over those without, based on human evaluation.}
\label{tab:human_evaluation}
\end{table}

\begin{figure*}[th]
    \centering
    \includegraphics[width=\textwidth]{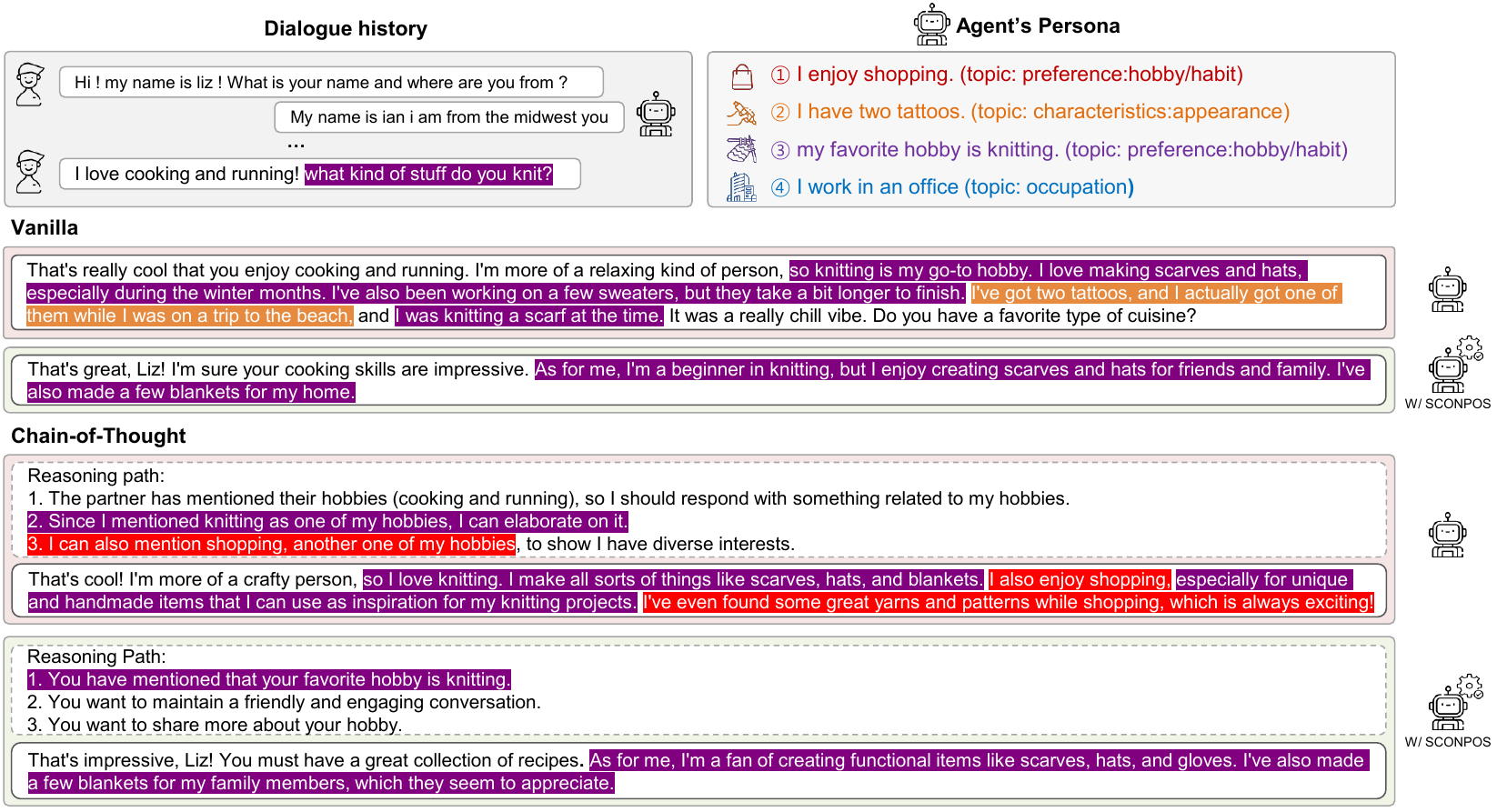}
    \caption{Case study of responses generated by Llama 3.1 on the PersonaChat dataset under vanilla and chain-of-thought settings. Highlighted text indicates persona-grounded expressions derived from the assigned persona attributes. The full version, including results of task decomposition and self-refinement, is presented in Figure~\ref{fig:case_study_full}. The quantitative impact of SCONPOS when combined with ICL-based methods is provided in Table~\ref{tab:CoT_steering_effect}.}
    \label{fig:Case_study_main_edit}
\end{figure*}

\paragraph{Human evaluation.}
To verify whether responses with mitigated overuse by SCONPOS are also perceived as more appropriate by humans, we conduct a human evaluation. As shown in Table \ref{tab:human_evaluation}, SCONPOS consistently achieves higher win rates in both persona appropriateness and dialogue coherence across all ICL-based methods and datasets. This confirms that the overuse mitigation effect of our method is valid not only in automatic evaluation metrics but also in human judgments.

\begin{figure}[h]
    \centering
    \includegraphics[width=\linewidth]{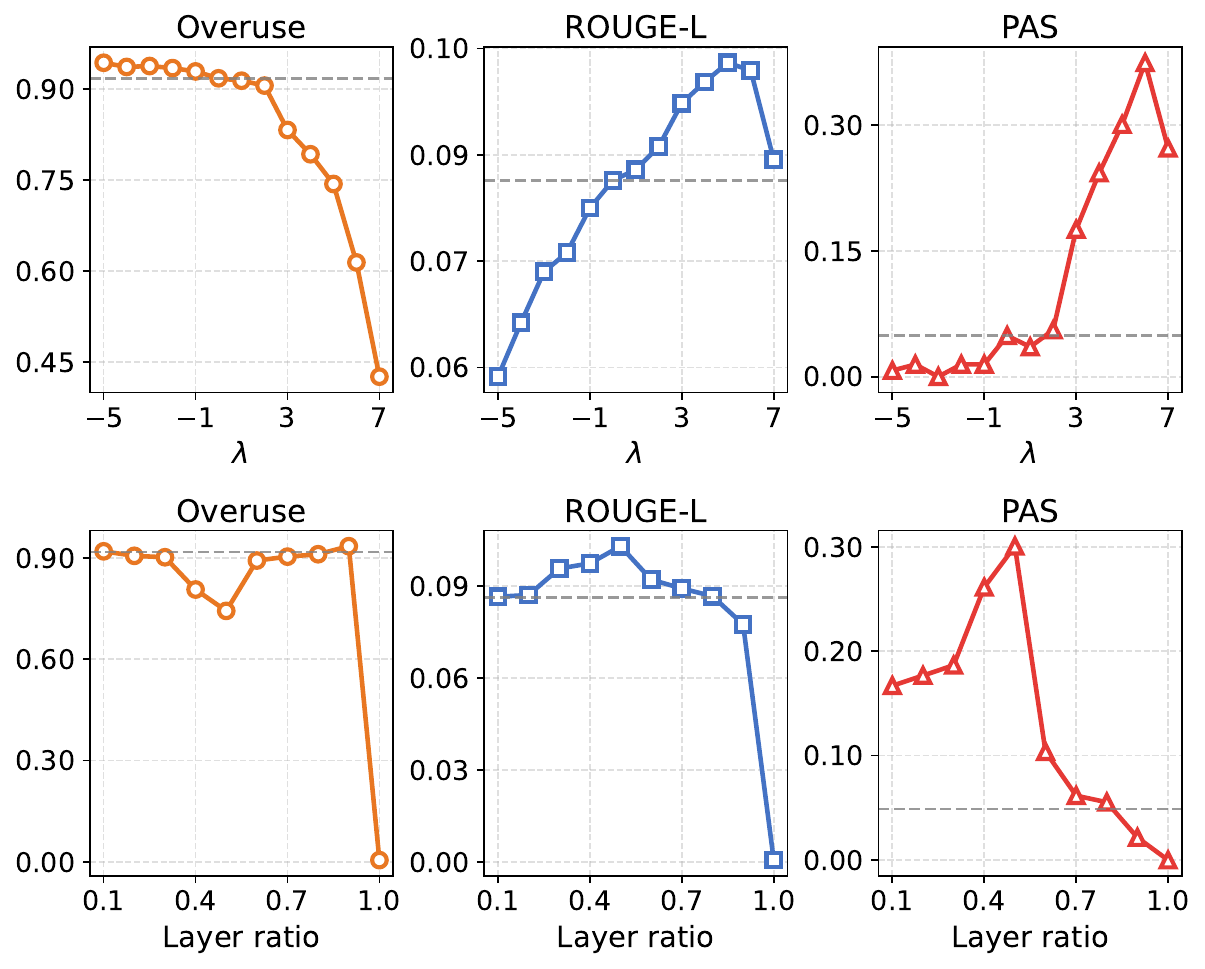}
    \caption{Performance results with varying suppression intensity $\lambda$ (top) and target layer ratio (bottom).}
    \label{fig:persona_llama_ablation}

\end{figure}

\paragraph{Hyperparameter analysis.}
To better understand the behavior of SCONPOS, we analyze how the suppression intensity $\lambda$ and target layer $l$ affect performance. Figure \ref{fig:persona_llama_ablation} shows the performance trends across suppression intensity $\lambda$ and target layer ratio, using Llama3.1 on PersonaChat. The layer ratio indicates the relative position of the target layer with respect to the total number of layers in the model. Results for all datasets and models are provided in Figure \ref{fig:ablation_combined}. 
\vspace{5pt}

As the suppression intensity $\lambda$ increases, Overuse decreases, while setting $\lambda$ to a negative value amplifies persona overuse. However, excessively large values of $\lambda$ can lead to abnormal activation values, disrupting the generation process. Regarding the target layer, although there are slight variations across models and datasets, persona use is most effectively controlled in the middle layers, with a layer ratio of 0.3 to 0.7.

\paragraph{Case study.} Figure \ref{fig:Case_study_main_edit} illustrates responses to a partner's question about knitting, the agent's favorite hobby. Without SCONPOS, the Vanilla setting produces unnatural responses by forcefully incorporating \textit{tattoo} which is irrelevant to \textit{knitting} into the dialogue. With SCONPOS, the response appropriately focuses on the partner's question. In the CoT setting, the tendency to mention \textit{shopping} (irrelevant persona attribute) already emerges in the reasoning path and carries over into the final response. With SCONPOS applied, however, the reasoning path itself is steered to reflect only knitting, which is relevant to the dialogue context. These results demonstrate that SCONPOS can not only suppress overuse in the final response but also alter the reasoning process itself by directly intervening at the prompt encoding stage where overuse-inducing bias is formed.

\begin{figure*}[t]
\centering
  \includegraphics[width=\textwidth]{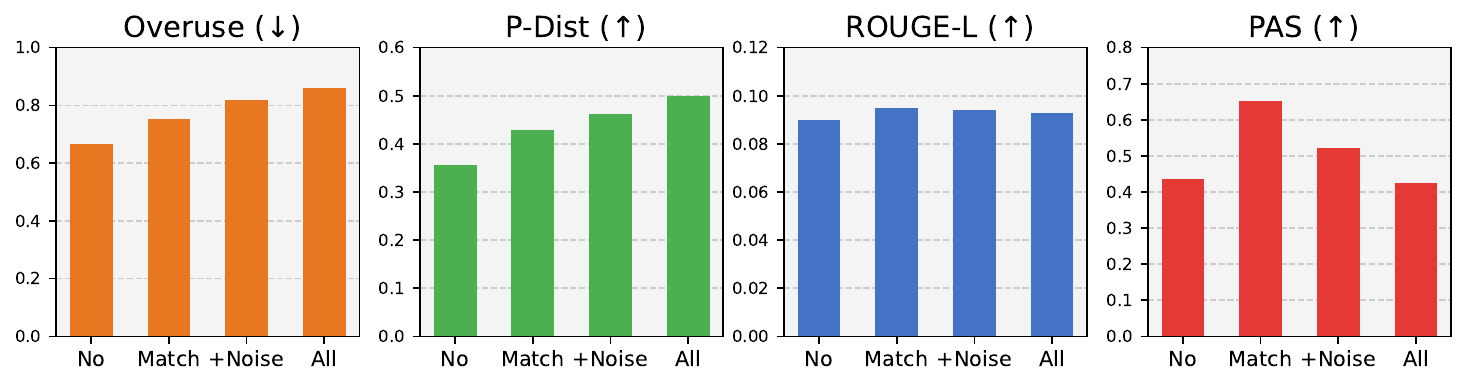}
\caption{Metric results on \textit{persona-relevant utterances} across different persona input settings. The values represent the average metric scores for responses generated by LLMs.}
  \label{fig:relevant_across_metrics}
\end{figure*}

\subsection{Validation of PAS}
We validate whether PAS measures the appropriateness of persona use, leveraging the observed tendency of LLMs to apply all given personas.

\paragraph{Can PAS detect both underuse and overuse?}
Figure~\ref{fig:relevant_across_metrics} presents the metric results for responses generated under different persona input settings for \textit{persona-relevant utterances}. PAS assigns the highest score to the Match setting, while assigning the lowest scores to the No and All settings. Furthermore, when comparing +Noise and All, PAS assigns a lower score to All, reflecting the more severe persona overuse. These results demonstrate that PAS exhibits \textbf{bidirectional sensitivity}, penalizing both underuse and overuse.

\begin{figure}[h]
\centering
  \includegraphics[width=\columnwidth]{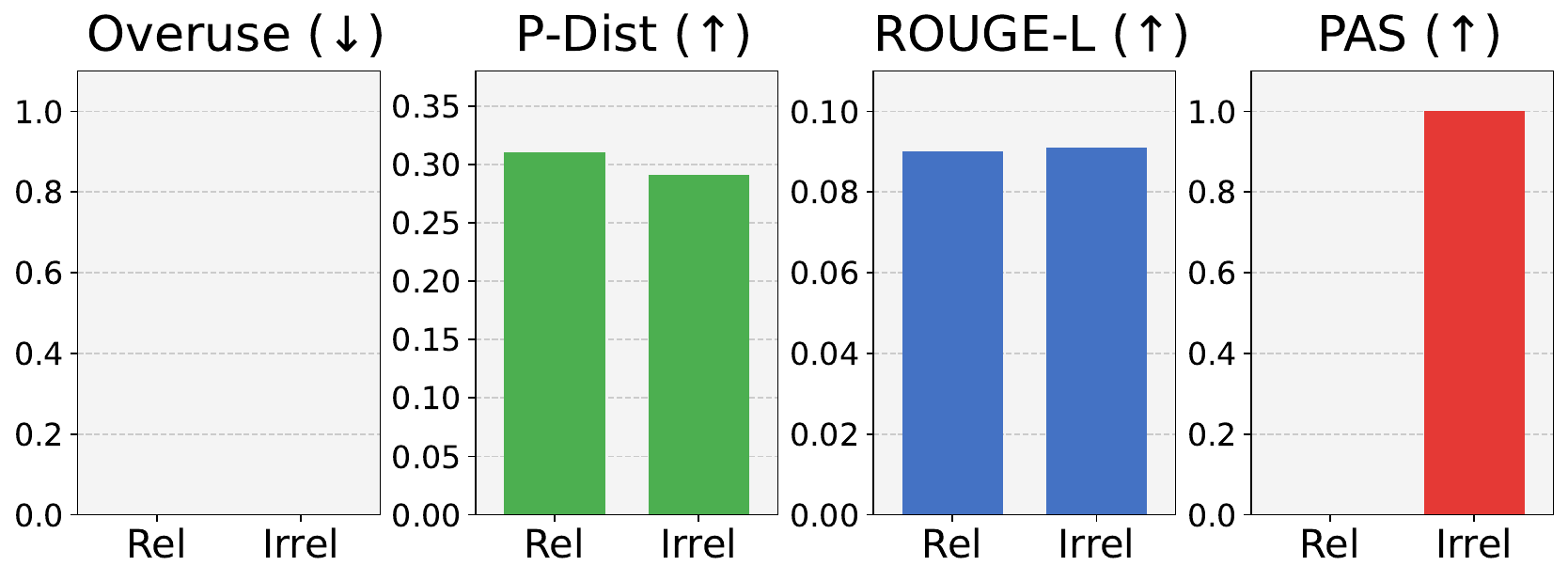}
  \caption{Metric results for responses under the \textbf{No} setting across \textit{persona-relevant} (Rel) and \textit{persona-irrelevant} (Irrel) utterances.}
  \label{fig:no_setting_relevant}
  \vspace{-15pt}
\end{figure}

\paragraph{Can PAS distinguish the appropriateness of persona use according to dialogue context?}
Figure~\ref{fig:no_setting_relevant} presents the metric results for responses generated under the No setting for both \textit{persona-relevant} and \textit{persona-irrelevant utterances}. Although both cases involve responses that do not reflect any persona attributes, the former constitutes underuse while the latter is considered appropriate, as excluding persona is contextually proper. Overuse assigned a score of 0 in both cases, and P-Dist and ROUGE-L showed only marginal differences. In contrast, PAS clearly distinguishes between the two cases. These results demonstrate that PAS exhibits \textbf{contextual sensitivity}, accurately evaluating the contextual appropriateness of persona use.
\begin{figure}[h]
\centering
  \includegraphics[width=\columnwidth]{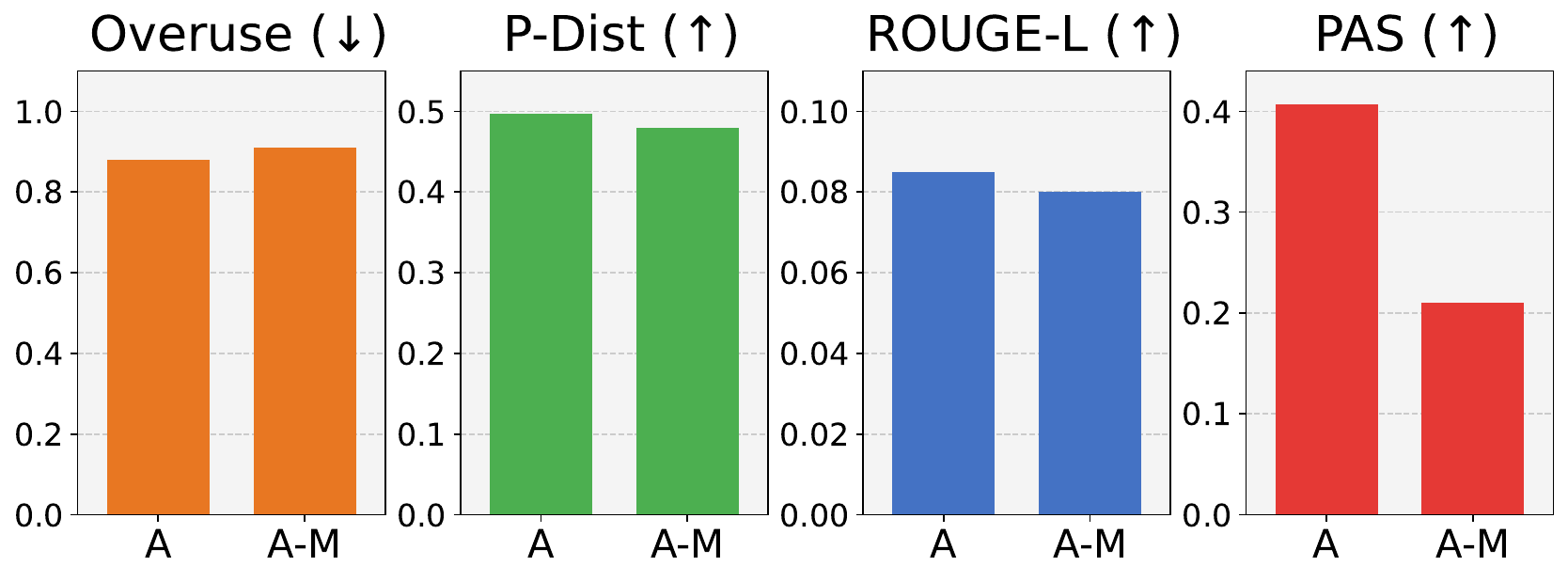}
  \caption{Metric results for responses generated under the \textbf{All} (A) and \textbf{All}-\textbf{Match} (A-M) settings. A-M denotes the setting where only contextually irrelevant persona attributes are provided.}
  \label{fig:4}
  \vspace{-10pt}
\end{figure}
\paragraph{Can PAS discriminate persona use when overuse occurs?}
Figure~\ref{fig:4} presents the metric results for responses generated under the \textbf{All} and \textbf{All}-\textbf{Match} settings. The \textbf{All} setting includes both contextually appropriate and irrelevant persona attributes, whereas the All-Match setting contains only irrelevant persona attributes. The results show that existing metrics have marginal differences between the two settings (Overuse: $-3.3\%$, P-Dist: $3.4\%$, ROUGE-L: $5.9\%$). In contrast, PAS shows a $48.4\%$ difference, clearly distinguishing between the two settings. These results demonstrate that PAS exhibits \textbf{discriminative sensitivity}, accurately capturing fine-grained differences in persona use even when overuse occurs.

\paragraph{Is PAS consistent with human judgments?}
Table~\ref{table:human_eval_pas} presents the Pearson correlation between human judgments of persona appropriateness and each metric. Among the compared metrics, PAS shows the highest correlation with statistical significance ($p < 0.05$). We also conduct an additional human evaluation (Appendix~\ref{app:human_eval}). As shown in Table~\ref{tab:human_eval_additional}, P-Dist shows negative correlations with human judgments. This suggests that the existing evaluation framework has been rewarding excessive persona use rather than contextually appropriate use. This highlights the need for a shift toward appropriateness-centered evaluation in PDG, and we believe that PAS is a first step in that direction.

\begin{table}[h]
\centering
\resizebox{\columnwidth}{!}{%
\begin{tabular}{@{}cccccc@{}}
\toprule
Metric & Overuse & C.Score & \multicolumn{1}{l}{P-Dist} & P-F1 & PAS \\ \midrule
Pearson's $r$ & -0.028 & 0.026 & 0.010 & -0.043 & \textbf{0.093} \\
$p$-value & 0.530 & 0.562 & 0.817 & 0.343 & 0.038 \\ \bottomrule
\end{tabular}%
}
\caption{Pearson correlation between evaluation metrics and human judgments of persona use appropriateness.}
\label{table:human_eval_pas}
\end{table}

\section{Conclusion}
In this study, we systematically analyzed the persona overuse problem in LLM-based PDG and proposed a method and evaluation metric to address it. Through analysis, we revealed that LLMs exhibit a tendency to incorporate all given persona attributes regardless of dialogue context, and based on this insight, we proposed SCONPOS, which mitigates overuse by modifying LLMs' internal representations in a response-free manner. Experimental results demonstrate that SCONPOS systematically mitigates overuse and confirm that overuse-inducing bias is formed at the prompt encoding stage and influences the reasoning process as a whole. These findings suggest that persona overuse is not a superficial phenomenon but stems from a fundamental bias within LLMs. Furthermore, we proposed and validated PAS to detect both overuse and underuse according to dialogue context, unlike existing metrics. We believe that this work serves as a step toward more contextually appropriate persona-based dialogue systems.

\section*{Limitations}
Despite the effectiveness of the proposed SCONPOS and PAS, we acknowledge the following limitations. First, the optimal $\lambda$ and target layer may vary depending on the model and dataset. Therefore, when applying SCONPOS to a new model or dataset, it is necessary to search for a new optimal configuration. Although we made substantial efforts to identify effective hyperparameters, the final settings we selected are not guaranteed to be globally optimal. Future work could address this by automating the search process or by developing adaptive methods that adjust hyperparameters.

Second, our control experiments (Table~\ref{tab:control_vector}) and cross-dataset transfer results (Table~\ref{tab:cross_dataset}) provide evidence that the steering effect stems from the overuse-inducing bias rather than properties such as attribute count or prompt length. However, completely disentangling this bias from all correlated factors in the representation space is a broader open problem in representation-level intervention.

Another limitation is that our experiments and analyses are limited to English datasets. We evaluated SCONPOS and PAS using PersonaChat and MBTI-S2Conv, both of which are English-based datasets. Therefore, it remains unclear whether our findings generalize to other languages. In particular, persona expression and dialogue styles can vary across languages. As a result, the persona overuse patterns observed in English datasets and the mitigation effect of SCONPOS may not necessarily transfer to cross-lingual settings. Future work should examine the cross-lingual generalizability of SCONPOS and PAS using persona-based dialogue datasets in diverse languages.

Lastly, when assessing appropriate persona use through proposed PAS, we define appropriateness based on the topical alignment between the model's response and the partner's last utterance. However, aligning with the topic of the last utterance does not always guarantee naturalness. In real-world dialogue, a new topic can be introduced even if it differs from the partner's last utterance. Therefore, our approach has inherent limitations in that view, without explicitly assessing broader conversational natural topic shifts. Nevertheless, PAS advances beyond prior metrics by comparing topic distributions, rather than solely counting the frequency of topic or persona attributes. Future work could explore extending the framework to incorporate topic transitions across the full dialogue context, rather than focusing only on the partner’s last utterance.

\section*{Ethics Statement}
This work aims to improve the controllability and evaluation of appropriate persona use in LLM-based PDG. While our goal is to make persona-based dialogue systems more contextually appropriate and reliable, we acknowledge the following potential risks.

First, SCONPOS could be misused. In this work, we use SCONPOS to reduce persona overuse in PDG. However, SCONPOS is a steering-based method that adjusts the internal representations of an LLM in a specific direction. If used maliciously, the steering direction could be reversed to encourage the model to overuse persona attributes that are irrelevant to the dialogue context. Furthermore, such steering could potentially be used to make an LLM excessively generate context-irrelevant knowledge, which may increase the risk of exposing sensitive or private information. In addition, if a malicious persona is assigned to the model, reversed or inappropriate steering could amplify harmful persona expressions and lead the model to generate more aggressive or inappropriate responses. These risks could become serious safety concerns in real-world PDG systems that directly interact with users.

Second, PAS should not be interpreted as an ethical or safety metric. PAS is designed to evaluate whether the persona used in a model response is appropriate with respect to the user's dialogue context. Therefore, a high PAS score does not necessarily indicate that the response is ethically safe or socially appropriate. For example, if the dialogue context itself is malicious, PAS may assign a high score to a response that uses a malicious persona in a contextually aligned way. Thus, PAS should not be used as a substitute for AI safety metrics that assess toxicity, harmfulness, bias, privacy risk, or other ethical concerns.

\section*{Acknowledgments}
This material is based upon work supported by the Air Force Office of Scientific Research under award number FA2386-23-1-4121, by the National Research Foundation of Korea (NRF) grant funded by the Korea government (MSIT) (RS-2024-00458720), and by the Institute of Information \& Communications Technology Planning \& Evaluation (IITP) grants funded by the Korean government (MSIT) (RS-2024-00439932, SW Starlab; No.RS-2020-II201336, Artificial Intelligence graduate school support (UNIST); No.RS-2021-II212068, Artificial Intelligence Innovation Hub; RS-2025-25442824, AI Star Fellowship Program (Ulsan National Institute of Science and Technology)). The authors used a generative AI tool for linguistic refinement and grammatical editing of the manuscript.

\bibliography{custom}

\clearpage

\appendix
\section{{Additional Details for Analysis}}
\label{sec:appendix}

Tables~\ref{append_table:irrelevant_analysis_agent_only_setting} and~\ref{append_table:relevant_analysis_agent_only_setting} present each model's results before averaging. 
Across the four models, we observed consistent patterns, confirming that the trends reported in Table~\ref{table:1} are robust.

\paragraph{Topic annotation.} For PersonaChat, we adopt the predefined topics and topic-assigned persona attributes from \citet{kim-etal-2024-panda}. For MBTI-S2Conv, we use the topics of each persona's attributes provided in the dataset itself. The topics of utterances and generated responses are annotated using GPT-4o, following the topic annotation framework of \citet{kim-etal-2024-panda}.

\subsection{Analysis Setup Details}
\paragraph{Model Parameter Setup}
We employed four instruction-tuned large language models: Llama-3.1-8B-Instruct, Qwen2.5-7B-Instruct, deepseek-llm-7b-chat, and gpt-4.1-2025-04-14. For the open-weight models, we used the official checkpoints with the default decoding parameters recommended by each provider (temperature = 1.0, top-p = 1.0). For GPT-4.1, we followed the default generation settings provided through the OpenAI API. The maximum number of output tokens was set to 256 for all tasks.

\paragraph{Topic Labeling Setup}
Based on the PANDA framework \cite{kim-etal-2024-panda}, we employed GPT-4o {(\texttt{gpt-4o-2024-08-06})} to label persona attributes and map them to predefined topics, which were used to measure persona overuse.
For PersonaChat, unlike the original PANDA setting, we independently mapped the topic of each partner's last utterance using GPT-4o.
This enables more accurate topic assignment by capturing not only explicit keywords but also implicit contextual meanings.
As a result, each last utterance in PersonaChat was assigned to a single most relevant topic, rather than multiple topics as in the original PANDA setting.
For MBTI-S2Conv, we used the utterance-level topic labels provided in the dataset.
Consequently, the excess-of-quantity criterion was defined as using more than one persona attribute beyond the single topic assigned to the partner's last utterance, making the evaluation setting more challenging.
Our proposed approach was designed to effectively mitigate overuse even under this stricter and more difficult configuration.

\paragraph{Evaluation of PAS setup.}
\label{pas_setup}
Both Figure~\ref{fig:no_setting_relevant} and Figure~\ref{fig:4} were generated using responses from Llama3.1 on the PersonaChat dataset under different persona input settings. Figure~\ref{fig:no_setting_relevant} corresponds to the \textbf{No} setting, representing the case where no persona attributes were provided. Among the total 2,100 generated responses, 504 did not include any persona information. Only these responses were used to evaluate the metric’s ability to distinguish between relevant utterances and irrelevant utterances. Although the \textbf{No} setting excludes predefined persona attributes, the model can still refer to persona-related information implicitly contained in the dialogue history. Therefore, we exclusively evaluate responses that show no trace of persona usage to ensure the purity of the \textbf{No} condition. Figure~\ref{fig:4} corresponds to the \textbf{All} and \textbf{All-Match} setting. Both settings generate responses to the same 724 persona-relevant utterances.

\subsection{Analysis Results with Both Agent and Partner Personas}
\label{anal_partner}
We further examine whether the trends observed in Section~\ref{sec:2} remain consistent when both the agent's persona and the partner's persona are provided to the model. Tables~\ref{append_table:irrelevant_analysis_partner_agent_setting} and~\ref{append_table:relevant_analysis_partner_agent_setting} report the results for persona-irrelevant and persona-relevant utterances, respectively. Overall, the results show patterns similar to those in the agent-only setting.

For both persona-irrelevant and persona-relevant utterances, overuse increases as more persona attributes are included in the input.
Even when the partner's persona is additionally provided, LLMs tend to incorporate persona information beyond what is contextually appropriate for the partner's current utterance.
This suggests that the presence of additional persona attributes strengthens the model's tendency to use persona information, rather than helping it selectively use only contextually relevant attributes.

The results also show that existing metrics have similar limitations in this setting.
Persona consistency metrics generally assign higher scores to settings with more persona attributes, indicating that they reward the amount of persona information expressed in the response rather than its contextual appropriateness.
Reference-based metrics show only small differences across persona settings and therefore do not reliably capture changes in persona use.
In addition, response length tends to increase as more persona attributes are provided, suggesting that additional persona information can encourage longer and more persona-heavy responses.

\section{Implementation Details} \label{append:sec:B}
\subsection{Prompt Templates}
\begin{figure*}
    \centering
    \includegraphics[width=\textwidth]{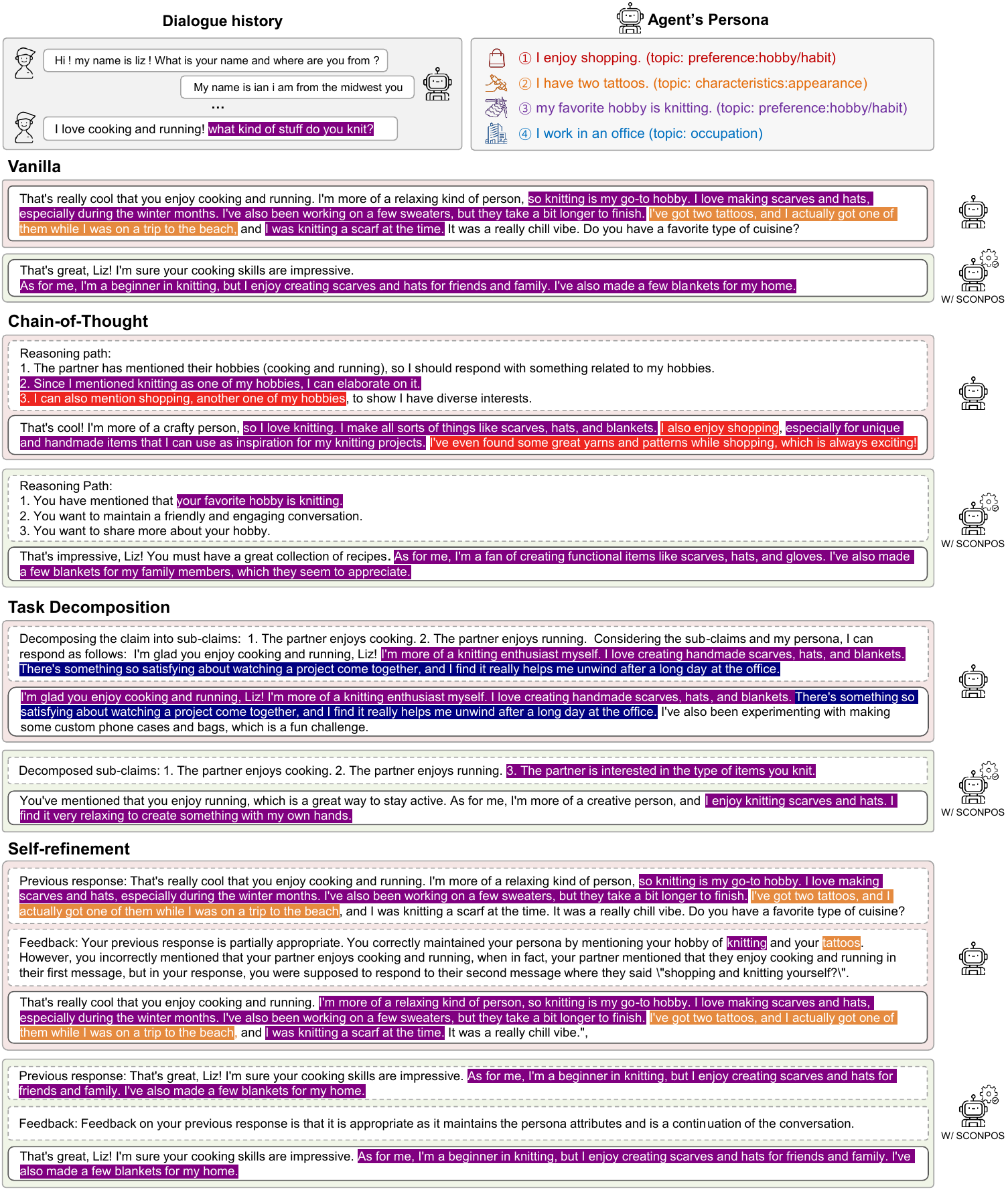}
    \caption{Case study of responses generated by Llama 3.1 on the PersonaChat 
            dataset under ICL-based settings.}
    \label{fig:case_study_full}
\end{figure*}

\begin{figure*}[h]
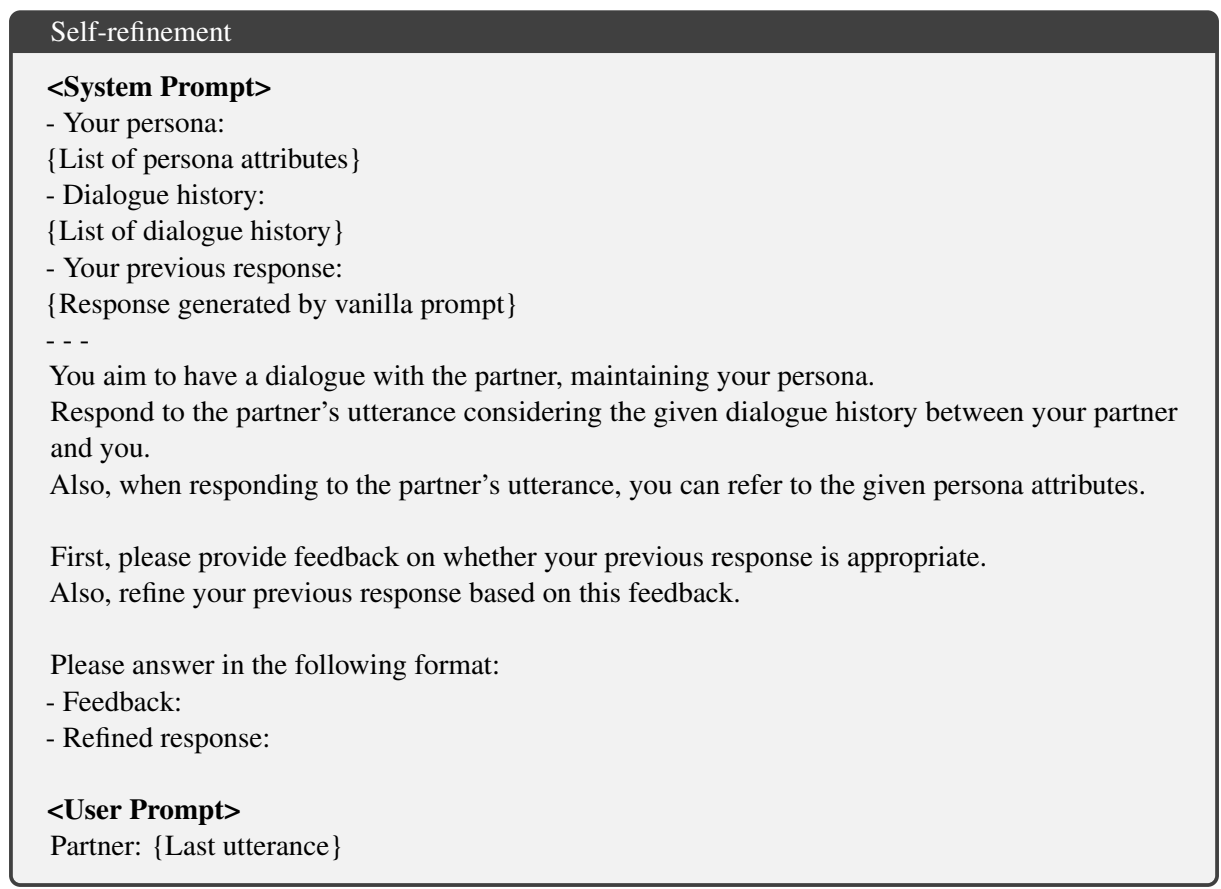

    \begin{tcolorbox}[title={Self-refinement}]

    \textbf{\textless System Prompt\textgreater} \\
    - Your persona:\\
    \{List of persona attributes\} \\
    - Dialogue history:\\
    \{List of dialogue history\} \\
    - Your previous response:\\
    \{Response generated by vanilla prompt\} \\
    - - - \\
    You aim to have a dialogue with the partner, maintaining your persona.\\
    Respond to the partner's utterance considering the given dialogue history between your partner and you.\\
    Also, when responding to the partner's utterance, you can refer to the given persona attributes.\\
    
    First, please provide feedback on whether your previous response is appropriate.\\
    Also, refine your previous response based on this feedback.\\
    
    Please answer in the following format:\\
    - Feedback:\\
    - Refined response:\\
        
    \textbf{\textless User Prompt\textgreater} \\
    Partner: \{Last utterance\}
    \end{tcolorbox}
    \caption{Prompt for Self-refinement.}
    \label{fig:prompt_sr}

\end{figure*}

For the experiments reported in Table~\ref{table:1}, we used the vanilla prompt templates shown in Figures~\ref{fig:prompt_vanilla} and~\ref{fig:prompt_vanilla_mbti} for PersonaChat and MBTI-S2Conv, respectively. 
For the case studies in Figure~\ref{fig:case_study_full} and the human evaluation in Table~\ref{tab:human_evaluation}, we used the prompt templates shown in Figures~\ref{fig:prompt_cot},~\ref{fig:prompt_td}, and~\ref{fig:prompt_sr}, corresponding to Chain-of-Thought (CoT), Task Decomposition, and Self-Refine, respectively. 
All prompt templates are based on those used in \citet{tu2023characterchat} and \citet{kim-etal-2024-panda}.

\subsection{Evaluation Metric Settings}

\paragraph{Overuse}
This metric first identifies the persona attributes grounded in the partner's utterance $u$ and the model response $\hat{y}$, and maps them into predefined persona topic categories. Overuse is then quantified by comparing the topic counts between $u$ and $\hat{y}$. 
For each topic $i$, the topic-level overuse score is computed as:
\begin{equation}
    ovs_i =
    \frac{|T^{\hat{y}}_i|}{\epsilon + |T^u_i|}
    \cdot \log(w_i),
\end{equation}
where $|T^u_i|$ and $|T^{\hat{y}}_i|$ denote the number of persona topics of type $i$ grounded in the partner's utterance and the model response, respectively, and $\epsilon$ is a small constant for numerical stability. The penalty weight $w_i$ is defined as:
\begin{equation}
w_i =
\begin{cases}
(x+1)e, & \text{if } |T^u_i| \neq 0 \text{ and } |T^u_i| < |T^{\hat{y}}_i|, \\
e^{x+1}, & \text{if } |T^u_i| = 0 \text{ and } |T^{\hat{y}}_i| > 0, \\
e, & \text{otherwise},
\end{cases}
\end{equation}
where $x = |T^{\hat{y}}_i| - |T^u_i|$ denotes the excess count, i.e., how many more times topic $i$ appears in the model response than in the partner's utterance. The final overuse score is obtained by averaging the topic-level scores over the union of topics appearing in $u$ and $\hat{y}$, followed by sigmoid normalization:
\begin{equation}
    \textrm{Overuse}(\hat{y} \mid u)
    =
    \sigma\left(
    \log
    \frac{
    \sum_{i=1}^{|T^u \cup T^{\hat{y}}|} ovs_i
    }{
    |T^u \cup T^{\hat{y}}|
    }
    \right).
\end{equation}

\paragraph{C.Score} 
C.Score measures the degree of entailment between each persona attribute and the generated response using a BERT-based natural language inference (NLI) model. 
We fine-tune \texttt{bert-large} \cite{devlin2019bert} on the DNLI dataset \cite{welleck-etal-2019-dialogue}. 
The fine-tuned model achieves an accuracy of 90.2\% on the DNLI test set.
We follow the original formulation of C.Score from \citet{madotto-etal-2019-personalizing}, defined as:

\begin{equation}
\label{eq:cscore}
\mathrm{NLI}(u, p_j) =
\begin{cases}
1  & \text{if } u \text{ entails } p_j,\\
0  & \text{if } u \text{ is independent of } p_j,\\
-1 & \text{if } u \text{ contradicts } p_j.
\end{cases}
\end{equation}
\begin{equation}
C(u) = \sum_{j=1}^{m} \mathrm{NLI}(u, p_j),
\end{equation}

where \(u\) denotes the generated utterance and \(p_j\) represents each persona attribute.

\paragraph{P-Distance}
P-Distance measures persona consistency by computing the semantic similarity between persona keywords and response keywords. For each persona keyword embedding $p_i$, we compute its cosine similarity with all response keyword embeddings $\{r_j\}_{j=1}^{m}$:
\begin{equation}
    M_i = [\mathrm{Sim}(p_i, r_1), \mathrm{Sim}(p_i, r_2), \ldots, \mathrm{Sim}(p_i, r_m)].
\end{equation}
The final P-Dist score is the average of the maximum similarity for each persona keyword:
\begin{equation}
    \mathrm{P\text{-}Dist}
    =
    \frac{1}{n}
    \sum_{i=1}^{n}
    \max_{1 \leq j \leq m} \mathrm{Sim}(p_i, r_j),
\end{equation}
where $n$ is the number of persona keywords.

\paragraph{P-F1}
P-F1 measures token-level overlap between a generated response and the given persona. Let $W_R$ and $W_P$ denote the sets of non-stopwords in the response and persona, respectively. Precision and recall are computed as:
\begin{equation}
    P = \frac{|W_R \cap W_P|}{|W_R|}, 
    \quad
    R = \frac{|W_R \cap W_P|}{|W_P|}.
\end{equation}
P-F1 is then defined as:
\begin{equation}
    \mathrm{P\text{-}F1}
    =
    \frac{2PR}{P+R}.
\end{equation}

\subsection{Human Evaluation Settings}
We recruited 10 volunteers who are undergraduate or graduate students and fluent in English. To validate the effectiveness of SCONPOS, annotators were presented with pairs of responses generated with and without steering for Vanilla and ICL-based methods. They were asked to select the superior response in terms of persona appropriateness and dialogue coherence, using the rubrics shown in Figures~\ref{fig:rubric_pa_method} and~\ref{fig:rubric_dc_method}, respectively. Each comparison set consisted of 50 pairs evaluated by three annotators, and the final judgment was determined by majority vote. 
We report the win rate. To validate PAS, annotators were asked to rate persona appropriateness on a 5-point Likert scale for different 50 samples from PersonaChat (total 500 samples). For the metric validation, we used the rubrics shown in Figures~\ref{fig:rubric_pa_PAS} and~\ref{fig:rubric_dc_PAS} for persona appropriateness and dialogue coherence, respectively.

\subsection{SCONPOS Settings}
We provide the detailed hyperparameter search and decoding settings used for SCONPOS and CAA.

\paragraph{Hyperparameter search range.}
We searched for the steering intensity $\lambda$ and the intervention target layer on the validation set. For $\lambda$, we searched values from 1 to 5 with an interval of 1. For the target layer, we searched layer ratios from 0.1 to 1.0 with an interval of 0.1, where the actual intervention layer was determined based on the ratio to the total number of layers in each model. SCONPOS and CAA used the same search range for both $\lambda$ and the target layer.

\paragraph{Selected hyperparameters.}
Table~\ref{tab:selected_hyperparams} reports the selected hyperparameters ($\lambda$ and target layer ratio) for SCONPOS and CAA across models and datasets.

\begin{table}[h]
\centering
\resizebox{\columnwidth}{!}{%
\begin{tabular}{@{}llcccc@{}}
\toprule
\multirow{2}{*}{Method} & \multirow{2}{*}{Model} & \multicolumn{2}{c}{PersonaChat} & \multicolumn{2}{c}{MBTI-S2Conv} \\ \cmidrule(lr){3-4} \cmidrule(l){5-6} 
 &  & $\lambda$ & Layer ratio & $\lambda$ & Layer ratio \\ \midrule
\multirow{3}{*}{SCONPOS} & Llama3.1 & 5.0 & 0.5 & 5.0 & 0.5 \\
 & Qwen2.5 & 5.0 & 0.7 & 5.0 & 0.5 \\
 & DeepSeek-chat & 2.0 & 0.5 & 2.0 & 0.5 \\ \midrule
\multirow{3}{*}{CAA} & Llama3.1 & 5.0 & 0.4 & 5.0 & 0.3 \\
 & Qwen2.5 & 5.0 & 0.6 & 5.0 & 0.7 \\
 & DeepSeek-chat & 2.0 & 0.5 & 2.0 & 1.0 \\ \bottomrule
\end{tabular}%
}
\caption{Selected hyperparameters (steering intensity $\lambda$ and target layer ratio) for SCONPOS and CAA across models and datasets.}
\label{tab:selected_hyperparams}
\end{table}

\paragraph{Decoding strategy.}
We used greedy decoding for all main experiments. That is, the final response generation for all methods, including SCONPOS and CAA, used the same decoding strategy for fair comparison. However, extracting the steering vector for CAA requires responses generated under the All and Match settings on the training set. Therefore, for the response generation step used only for CAA vector extraction, we used sampling-based decoding. Specifically, responses under the All and Match settings were generated with temperature 1.0 and top-p 1.0. These responses were used only for extracting the CAA steering vector, while greedy decoding was used for final response generation in evaluation.

\section{Additional Experimental Results}
\subsection{Additional Human Evaluation for PAS}
\label{app:human_eval}
To further examine the alignment between PAS and human judgments, we conducted an additional human evaluation with a refined protocol via Amazon Mechanical Turk. In the original protocol (Appendix~\ref{append:sec:B}), each of the 10 annotators rated a distinct set of 50 samples, yielding 500 non-overlapping samples in total but only a single rating per sample, so inter-annotator agreement could not be measured. In the new protocol, we annotated 100 response samples, each rated by three workers on a 3-point Likert scale, and aggregated the scores at the sample level. To ensure annotation quality, annotators were required to have a HIT Approval Rate above 95\%, more than 1,000 approved HITs, and residency in an English-speaking country (US, UK, Canada, Australia, or Singapore). We paid 30 cents per annotated sample, corresponding to a reasonable hourly rate based on the average completion time. The inter-annotator agreement, measured by Krippendorff's $\alpha$, was 0.155, indicating low agreement. This suggests that persona appropriateness is an inherently subjective concept on which even human annotators struggle to reach consensus.

\begin{table}[h]
\centering
\resizebox{\columnwidth}{!}{%
\begin{tabular}{@{}cccccc@{}}
\toprule
Metric & Overuse & C.Score & \multicolumn{1}{l}{P-Dist} & P-F1 & PAS \\ \midrule
Pearson's $r$ & {\ul -0.197} & -0.054 & {\ul -0.235} & 0.028 & \textbf{0.155} \\
$p$-value & 0.050 & 0.595 & 0.019 & 0.779 & 0.123 \\ \bottomrule
\end{tabular}%
}
\caption{Pearson correlation between evaluation metrics and human judgments of persona appropriateness under the redesigned evaluation protocol. Underline indicates a significant correlation ($p<0.05$); bold indicates the highest correlation.}
\label{tab:human_eval_additional}
\end{table}

\begin{table*}[t]
\centering
\resizebox{0.85\textwidth}{!}{%
\begin{tabular}{@{}lcccccccc@{}}
\toprule
\multirow{2}{*}{\begin{tabular}[c]{@{}l@{}}Train data ratio\end{tabular}} & \multicolumn{3}{c}{PersonaChat} &  & \multicolumn{3}{c}{MBTI-S2Conv} \\ \cmidrule(lr){2-4} \cmidrule(l){6-8} 
 & Overuse ($\downarrow$) & ROUGE-L ($\uparrow$) & PAS ($\uparrow$) &  & Overuse ($\downarrow$) & ROUGE-L ($\uparrow$) & PAS ($\uparrow$) \\ \midrule
Vanilla & 0.858 & 0.096 & 0.157 &  & 0.805 & 0.135 & 0.385 \\
5\%   & 0.717 & 0.104 & 0.306 &  & \textbf{0.700} & \textbf{0.143} & 0.513 \\
10\%  & 0.700 & 0.103 & 0.335 &  & 0.709 & 0.141 & 0.513 \\
25\%  & 0.673 & \textbf{0.106} & 0.362 &  & 0.710 & 0.141 & \textbf{0.517} \\
50\%  & 0.672 & \textbf{0.106} & 0.358 &  & 0.705 & 0.142 & \textbf{0.517} \\
100\% & \textbf{0.652} & \textbf{0.106} & \textbf{0.377} &  & 0.708 & 0.142 & 0.508 \\ \bottomrule
\end{tabular}%
}
\caption{Performance of SCONPOS with varying proportions of training data. Vanilla indicates the baseline setting without applying SCONPOS. Scores are averaged across Llama3.1, Qwen2.5, and DeepSeek-chat.}
\label{tab:data_efficiency}
\end{table*}

\begin{table*}[h]
\centering
\resizebox{0.85\textwidth}{!}{%
\begin{tabular}{@{}lcccccccc@{}}
\toprule
\multirow{2}{*}{Method} & \multicolumn{3}{c}{PersonaChat} &  & \multicolumn{3}{c}{MBTI-S2Conv} \\ \cmidrule(lr){2-4} \cmidrule(l){6-8} 
 & Overuse ($\downarrow$) & ROUGE-L ($\uparrow$) & PAS ($\uparrow$) &  & Overuse ($\downarrow$) & ROUGE-L ($\uparrow$) & PAS ($\uparrow$) \\ \midrule
Vanilla        & 0.858 & 0.096 & 0.157 &  & 0.805 & 0.135 & 0.385 \\
One layer (SCONPOS)   & 0.652 & \textbf{0.106} & \textbf{0.377} &  & 0.708 & 0.142 & \textbf{0.508} \\
Mid 3 layers   & \textbf{0.591} & 0.105 & 0.352 &  & \textbf{0.650} & \textbf{0.149} & 0.490 \\
All layers     & 0.690 & 0.093 & 0.289 &  & 0.663 & 0.127 & 0.468 \\ \bottomrule
\end{tabular}%
}
\caption{Performance comparison of applying the steering vector to a single target layer (SCONPOS), three middle layers, and all layers. Scores are averaged across Llama3.1, Qwen2.5, and DeepSeek-chat.}
\label{tab:num_layers_performance}
\end{table*}

Table~\ref{tab:human_eval_additional} reports the Pearson correlation between each automatic metric and human judgments under the redesigned protocol. Combining these results with the original protocol (Table~\ref{table:human_eval_pas}), PAS is the only metric that shows a consistently positive correlation with human appropriateness judgments across both protocols. Under the redesigned protocol, P-Dist shows a significant negative correlation ($r=-0.235$, $p=0.019$), which is consistent with our observation in Section~\ref{sec:2} that persona consistency metrics do not reflect contextual appropriateness. Overuse also shows a negative correlation ($r=-0.197$, $p=0.050$), while C.Score and P-F1 show no discernible relationship with human judgments in either evaluation.

\subsection{Data efficiency}
To examine the sample efficiency of steering vector extraction, we conducted additional experiments using varying proportions of training data to extract the steering vector. As shown in Table \ref{tab:data_efficiency}, SCONPOS achieves comparable performance with only a small fraction of the training data. Even with just 5\% of the data, Overuse and PAS improve substantially over Vanilla. On MBTI-S2Conv, the 5--10\% setting performs comparably to the 100\% setting, and on PersonaChat, performance largely saturates at 25\%. These results show that although SCONPOS requires in-domain samples, the required amount is relatively small, posing a limited practical burden.

\subsection{Multi-layer intervention}
We conducted a comparison experiment by applying the steering vector to a single layer, three middle layers, and all layers. As shown in Table \ref{tab:num_layers_performance}, applying the vector to three middle layers yields the lowest Overuse, but PAS is lower than with a single layer. Applying to all layers further degrades not only PAS but also ROUGE-L. These results show that intervening across multiple layers simultaneously leads to performance degradation rather than cumulative improvement.

We also measured the time cost of each setting for an efficiency comparison. As shown in Table \ref{tab:num_layers_cost}, vector extraction time increases slightly as the number of intervened layers grows. Moreover, multi-layer intervention combinatorially expands the search space for selecting which layer combination to use, increasing the hyperparameter optimization burden. In summary, single-layer intervention achieves the best performance in PAS while incurring the smallest extraction cost and search space, confirming it as a reasonable design choice in terms of both performance and practicality.

\begin{table}[h]
\centering
\resizebox{\columnwidth}{!}{%
\begin{tabular}{@{}lcc@{}}
\toprule
Method & \begin{tabular}[c]{@{}c@{}}Vector extraction time \\ (min)\end{tabular} & \begin{tabular}[c]{@{}c@{}}Inference time\\ (sec/it)\end{tabular} \\ \midrule
One layer (SCONPOS) & 56.9 & 4.20 \\
Mid 3 layers        & 57.7 & 4.52 \\
All layers          & 59.0 & 5.22 \\ \bottomrule
\end{tabular}%
}
\caption{Time cost comparison of applying the steering vector to a single target layer (SCONPOS), three middle layers, and all layers with the MBTI-S2Conv dataset.}
\label{tab:num_layers_cost}
\end{table}

\section{Related Works}
\subsection{Persona-based Dialogue Generation}
Early work framed PDG as the generation of conditioning responses on a set of persona attributes \cite{li-etal-2016-persona}, later popularized by PersonaChat \cite{zhang-etal-2018-personalizing}. 
Earlier studies mainly focused on improving persona consistency.
Prior works utilized unlikelihood training \cite{song-etal-2021-bob, li-etal-2020-dont} or reinforcement learning \cite{song2020generating, shea2023building, takayama-etal-2025-persona} to discourage non-entailed persona use.
However, these methods rely on large amounts of data, which makes them costly and less scalable to new domains or unseen personas.
With larger pretrained model, prompt-based approaches, such as ICL, have achieved strong persona consistency without additional parameter updates \cite{madotto2021few, zheng2021exploring, xu-etal-2023-towards-zero, ni-etal-2023-multi}. 
However, recent studies reveal a new problem persona \textit{overuse} where models use persona even when off-topic or in excessive quantity \cite{kim-etal-2024-panda}. 
This work investigates why \textit{overuse} arises under different persona input conditions and proposes a representation-level intervention that suppresses the use of irrelevant persona attributes.

\subsection{Evaluation Metric in PDG}
Evaluation in PDG task can be broadly categorized into two perspectives: fluency and persona consistency.
Fluency measures how close a generated response is to a human reference. It is typically evaluated using automatic overlap-based metrics such as BLEU and ROUGE-L, which quantify lexical overlap between the generated and reference responses without requiring any additional trained model.
Embedding-based metrics \cite{zhang2019bertscore, zhao-etal-2019-moverscore}, in contrast, assess semantic similarity by computing embedding distances using pretrained contextual encoders such as BERT \cite{devlin2019bert}.
For the PDG task, evaluation focuses on persona consistency, which measures how well the response is grounded in the given persona.
Prior works estimated this groundedness by measuring the similarity between the generated response and persona attributes, either through word-level overlap \cite{lian2019learning, song2019exploiting} or vector-based similarity using BERT models trained on the Dialogue NLI dataset \cite{welleck-etal-2019-dialogue, madotto-etal-2019-personalizing, cho-etal-2022-personalized, song2020generating}.
However, these metrics mainly focus on measuring how much predefined personas are reflected in the response, rather than whether its usage is appropriate to the partner’s utterance.

\begin{table}[!b]
\centering
\resizebox{\columnwidth}{!}{%
\begin{tabular}{@{}llcc@{}}
\toprule
Model & Cost & CAA & SCONPOS \\ \midrule
\multirow{5}{*}{Llama3.1}
 & Total extraction time (min) & 254.2 & 56.9 \\
 & \quad Response generation time & 180.5 & - \\
 & \quad Vector extraction time & 73.7 & 56.9 \\
 & Inference time (sec/it) & 4.32 & 4.20 \\
 & Storage space (KB) & 20.0 & 20.0 \\ \midrule
\multirow{5}{*}{Qwen2.5}
 & Total extraction time (min) & 199.4 & 52.3 \\
 & \quad Response generation time & 130.4 & - \\
 & \quad Vector extraction time & 69.0 & 52.3 \\
 & Inference time (sec/it) & 6.12 & 5.80 \\
 & Storage space (KB) & 16.8 & 16.8 \\ \midrule
\multirow{5}{*}{DeepSeek-chat}
 & Total extraction time (min) & 223.6 & 51.0 \\
 & \quad Response generation time & 157.4 & - \\
 & \quad Vector extraction time & 66.2 & 51.0 \\
 & Inference time (sec/it) & 5.23 & 4.80 \\
 & Storage space (KB) & 18.8 & 18.8 \\ \midrule\midrule
\multirow{5}{*}{Average}
 & Total extraction time (min) & 225.7 & 53.4 \\
 & \quad Response generation time & 156.1 & - \\
 & \quad Vector extraction time & 69.6 & 53.4 \\
 & Inference time (sec/it) & 5.22 & 4.93 \\
 & Storage space (KB) & 18.5 & 18.5 \\ \bottomrule
\end{tabular}%
}
\caption{Time and storage cost comparison between CAA and SCONPOS across models with the MBTI-S2Conv dataset.}
\label{tab:cost_efficiency_full}
\end{table}

\section{Licenses and Use of Artifacts}
Table~\ref{tab:artifacts} lists artifacts used in this work, including datasets, models, and software libraries, with their licenses and links. Our use of the artifacts is consistent with their licenses and intended use. We do not redistribute any artifact in a manner inconsistent with its original license or access conditions, and any outputs or derived materials from this work are intended only for research use.

\begin{table*}[]
\centering
\setlength{\tabcolsep}{3pt}
\resizebox{\textwidth}{!}{%
\begin{tabular}{@{}cclccccccccccccccccccc@{}}
\toprule
\multirow{3}{*}{Model} & \multirow{3}{*}{\begin{tabular}[c]{@{}c@{}}Persona\\ setting\end{tabular}} & \multicolumn{1}{c}{} & \multicolumn{9}{c}{PersonaChat} &  & \multicolumn{9}{c}{MBTI-S2Conv} \\ \cmidrule(lr){4-12} \cmidrule(l){14-22} 
 &  &  & \multirow{2}{*}{\textbf{Overuse} ($\downarrow$)} &  & \multicolumn{3}{c}{Consistency ($\uparrow$)} &  & \multicolumn{3}{c}{Ref. Similarity ($\uparrow$)} &  & \multirow{2}{*}{\textbf{Overuse} ($\downarrow$)} &  & \multicolumn{3}{c}{Consistency ($\uparrow$)} & \textbf{} & \multicolumn{3}{c}{Ref. Similarity ($\uparrow$)} \\ \cmidrule(lr){6-8} \cmidrule(lr){10-12} \cmidrule(lr){16-18} \cmidrule(l){20-22} 
 &  &  &  &  & C.Score & P-Dist & P-F1 &  & BLEU & R-L & B.Score &  &  &  &  & P-Dist & P-F1 &  & BLEU & R-L & B.Score \\ \midrule
\multirow{3}{*}{Llama3.1} & \textbf{No} &  & \textbf{0.651} &  & 0.236 & 0.352 & 0.056 &  & 0.040 & \textbf{0.087} & \textbf{0.832} &  & \textbf{0.360} &  &  & 0.249 & 0.089 &  & \textbf{0.122} & \textbf{0.138} & \textbf{0.845} \\
 & \textbf{+Noise} &  & 0.881 &  & 0.622 & 0.411 & 0.066 &  & \textbf{0.041} & 0.084 & 0.831 &  & 0.627 &  &  & 0.285 & 0.101 &  & 0.105 & 0.124 & 0.842 \\
 & \textbf{All} &  & 0.940 &  & \textbf{0.913} & \textbf{0.493} & \textbf{0.071} &  & 0.039 & 0.083 & 0.829 &  & 0.741 &  &  & \textbf{0.298} & \textbf{0.104} &  & 0.100 & 0.123 & 0.841 \\ \midrule
\multirow{3}{*}{Qwen2.5} & \textbf{No} &  & \textbf{0.574} &  & 0.049 & 0.308 & 0.051 &  & \textbf{0.047} & \textbf{0.109} & \textbf{0.837} &  & \textbf{0.427} &  &  & 0.268 & 0.096 &  & \textbf{0.138} & \textbf{0.156} & \textbf{0.854} \\
 & \textbf{+Noise} &  & 0.841 &  & 0.466 & 0.381 & 0.066 &  & \textbf{0.047} & 0.103 & 0.836 &  & 0.609 &  &  & \textbf{0.281} & \textbf{0.097} &  & 0.125 & 0.144 & 0.852 \\
 & \textbf{All} &  & 0.918 &  & \textbf{0.845} & \textbf{0.468} & \textbf{0.073} &  & 0.044 & 0.096 & 0.833 &  & 0.650 &  &  & 0.278 & 0.093 &  & 0.125 & 0.149 & 0.852 \\ \midrule
\multirow{3}{*}{DeepSeek-chat} & \textbf{No} &  & \textbf{0.506} &  & 0.085 & 0.333 & 0.052 &  & 0.034 & 0.081 & 0.827 &  & \textbf{0.306} &  &  & 0.208 & 0.078 &  & \textbf{0.156} & \textbf{0.185} & \textbf{0.854} \\
 & \textbf{+Noise} &  & 0.682 & \multicolumn{1}{l}{} & 0.432 & 0.380 & 0.063 & \multicolumn{1}{l}{} & \textbf{0.037} & \textbf{0.089} & \textbf{0.831} & \multicolumn{1}{l}{} & 0.401 & \multicolumn{1}{l}{} & \multicolumn{1}{l}{} & 0.228 & 0.083 & \multicolumn{1}{l}{} & 0.150 & 0.172 & 0.852 \\
 & \textbf{All} & \multicolumn{1}{c}{\textbf{}} & 0.801 &  & \textbf{1.034} & \textbf{0.518} & \textbf{0.073} &  & 0.035 & 0.081 & 0.829 &  & 0.407 &  &  & \textbf{0.236} & \textbf{0.090} & \textbf{} & 0.142 & 0.166 & 0.853 \\ \midrule
\multirow{3}{*}{GPT4.1} & \textbf{No} & \textbf{} & \textbf{0.661} &  & 0.080 & 0.338 & 0.050 & \textbf{} & \textbf{0.049} & \textbf{0.110} & \textbf{0.835} &  & \textbf{0.344} &  &  & 0.235 & 0.084 &  & \textbf{0.135} & \textbf{0.163} & \textbf{0.851} \\
 & \textbf{+Noise} &  & 0.887 & \multicolumn{1}{l}{} & 0.549 & 0.425 & 0.071 & \multicolumn{1}{l}{} & \textbf{0.049} & 0.106 & \textbf{0.835} & \multicolumn{1}{l}{} & 0.527 & \multicolumn{1}{l}{} & \multicolumn{1}{l}{} & 0.263 & 0.093 & \multicolumn{1}{l}{} & 0.122 & 0.146 & 0.849 \\
 & \textbf{All} & \multicolumn{1}{c}{\textbf{}} & 0.938 &  & \textbf{0.883} & \textbf{0.506} & \textbf{0.078} &  & \textbf{0.049} & 0.104 & 0.834 &  & 0.616 &  &  & \textbf{0.273} & \textbf{0.096} & \textbf{} & 0.116 & 0.146 & 0.848 \\ \bottomrule
\end{tabular}%
}
\caption{Persona-irrelevant utterances in Agent-only setting.}
\label{append_table:irrelevant_analysis_agent_only_setting}
\end{table*}

\begin{table*}[]
\centering
\setlength{\tabcolsep}{3pt}
\resizebox{\textwidth}{!}{%
\begin{tabular}{@{}cclccccccccccccccccccc@{}}
\toprule
\multirow{3}{*}{Model} & \multirow{3}{*}{\begin{tabular}[c]{@{}c@{}}Persona\\ setting\end{tabular}} & \multicolumn{1}{c}{} & \multicolumn{9}{c}{PersonaChat} &  & \multicolumn{9}{c}{MBTI-S2Conv} \\ \cmidrule(lr){4-12} \cmidrule(l){14-22} 
 &  &  & \multirow{2}{*}{\textbf{Overuse} ($\downarrow$)} &  & \multicolumn{3}{c}{Consistency ($\uparrow$)} &  & \multicolumn{3}{c}{Ref. Similarity ($\uparrow$)} &  & \multirow{2}{*}{\textbf{Overuse} ($\downarrow$)} &  & \multicolumn{3}{c}{Consistency ($\uparrow$)} & \textbf{} & \multicolumn{3}{c}{Ref. Similarity ($\uparrow$)} \\ \cmidrule(lr){6-8} \cmidrule(lr){10-12} \cmidrule(lr){16-18} \cmidrule(l){20-22} 
 &  &  &  &  & C.Score & P-Dist & P-F1 &  & BLEU & R-L & B.Score &  &  &  &  & P-Dist & P-F1 &  & BLEU & R-L & B.Score \\ \midrule
\multirow{4}{*}{Llama3.1} & \textbf{No} &  & \textbf{0.730} &  & 0.173 & 0.374 & 0.056 &  & 0.041 & 0.086 & \textbf{0.833} &  & \textbf{0.861} &  &  & 0.345 & \textbf{0.121} &  & \textbf{0.113} & \textbf{0.122} & \textbf{0.848} \\
 & \textbf{Match} &  & 0.761 &  & 0.649 & 0.429 & 0.066 &  & \textbf{0.043} & \textbf{0.088} & \textbf{0.833} &  & 0.867 &  &  & 0.364 & \textbf{0.121} &  & 0.107 & 0.120 & 0.847 \\
 & \textbf{+Noise} &  & 0.820 &  & 0.758 & 0.462 & 0.067 &  & 0.042 & \textbf{0.088} & \textbf{0.833} &  & 0.893 &  &  & 0.370 & \textbf{0.121} &  & 0.103 & 0.118 & 0.846 \\
 & \textbf{All} &  & 0.881 &  & \textbf{0.884} & \textbf{0.497} & \textbf{0.071} &  & 0.041 & 0.085 & 0.831 &  & 0.913 &  &  & \textbf{0.380} & \textbf{0.121} &  & 0.099 & 0.116 & 0.846 \\ \midrule
\multirow{4}{*}{Qwen2.5} & \textbf{No} &  & \textbf{0.641} &  & -0.011 & 0.333 & 0.055 &  & 0.047 & 0.097 & 0.837 &  & \textbf{0.857} &  &  & 0.337 & 0.112 &  & \textbf{0.112} & \textbf{0.122} & \textbf{0.856} \\
 & \textbf{Match} &  & 0.761 & \multicolumn{1}{l}{} & 0.607 & 0.417 & 0.073 & \multicolumn{1}{l}{} & \textbf{0.048} & \textbf{0.102} & \textbf{0.838} & \multicolumn{1}{l}{} & 0.865 & \multicolumn{1}{l}{} & \multicolumn{1}{l}{} & 0.345 & 0.112 & \multicolumn{1}{l}{} & 0.109 & \textbf{0.122} & \textbf{0.856} \\
 & \textbf{+Noise} &  & 0.827 &  & 0.764 & 0.452 & 0.075 &  & 0.047 & 0.098 & 0.836 &  & 0.878 &  &  & 0.347 & \textbf{0.113} &  & 0.109 & 0.121 & 0.855 \\
 & \textbf{All} &  & 0.878 &  & \textbf{0.822} & \textbf{0.482} & \textbf{0.075} &  & 0.047 & 0.098 & 0.834 &  & 0.890 &  &  & \textbf{0.351} & 0.110 &  & 0.102 & 0.118 & 0.854 \\ \midrule
\multirow{4}{*}{DeepSeek-chat} & \textbf{No} &  & \textbf{0.566} &  & 0.071 & 0.355 & 0.047 &  & 0.037 & 0.081 & 0.831 &  & \textbf{0.824} &  &  & 0.344 & 0.116 &  & 0.113 & 0.132 & 0.859 \\
 & \textbf{Match} &  & 0.682 & \multicolumn{1}{l}{} & 0.483 & 0.412 & 0.065 & \multicolumn{1}{l}{} & \textbf{0.042} & \textbf{0.090} & \textbf{0.834} & \multicolumn{1}{l}{} & 0.849 & \multicolumn{1}{l}{} & \multicolumn{1}{l}{} & \textbf{0.350} & \textbf{0.120} & \multicolumn{1}{l}{} & 0.114 & 0.133 & 0.858 \\
 & \textbf{+Noise} &  & 0.752 & \multicolumn{1}{l}{} & 0.629 & 0.445 & 0.071 & \multicolumn{1}{l}{} & 0.041 & 0.089 & 0.833 & \multicolumn{1}{l}{} & 0.842 & \multicolumn{1}{l}{} & \multicolumn{1}{l}{} & \textbf{0.350} & \textbf{0.120} & \multicolumn{1}{l}{} & 0.114 & 0.132 & 0.859 \\
 & \textbf{All} & \multicolumn{1}{c}{\textbf{}} & 0.777 &  & \textbf{0.835} & \textbf{0.497} & \textbf{0.073} &  & 0.041 & 0.089 & 0.833 &  & 0.842 &  &  & 0.344 & \textbf{0.120} & \textbf{} & \textbf{0.121} & \textbf{0.135} & \textbf{0.860} \\ \midrule
\multirow{4}{*}{GPT4.1} & \textbf{No} & \textbf{} & \textbf{0.725} &  & 0.048 & 0.363 & 0.053 & \textbf{} & \textbf{0.050} & 0.097 & \textbf{0.835} &  & \textbf{0.889} &  &  & 0.362 & \textbf{0.113} &  & \textbf{0.094} & \textbf{0.111} & \textbf{0.851} \\
 & \textbf{Match} &  & 0.811 & \multicolumn{1}{l}{} & 0.747 & 0.455 & 0.077 & \multicolumn{1}{l}{} & \textbf{0.050} & 0.101 & \textbf{0.835} & \multicolumn{1}{l}{} & 0.885 & \multicolumn{1}{l}{} & \multicolumn{1}{l}{} & 0.376 & 0.112 & \multicolumn{1}{l}{} & 0.091 & 0.107 & 0.848 \\
 & \textbf{+Noise} &  & 0.869 & \multicolumn{1}{l}{} & 0.863 & 0.492 & 0.078 & \multicolumn{1}{l}{} & \textbf{0.050} & \textbf{0.102} & \textbf{0.835} & \multicolumn{1}{l}{} & 0.900 & \multicolumn{1}{l}{} & \multicolumn{1}{l}{} & 0.382 & 0.111 & \multicolumn{1}{l}{} & 0.087 & 0.104 & 0.846 \\
 & \textbf{All} & \multicolumn{1}{c}{\textbf{}} & 0.900 &  & \textbf{0.932} & \textbf{0.521} & \textbf{0.081} &  & \textbf{0.050} & 0.100 & 0.834 &  & 0.913 &  &  & \textbf{0.391} & 0.110 & \textbf{} & 0.082 & 0.101 & 0.843 \\ \bottomrule
\end{tabular}%
}
\caption{Persona-relevant utterances in Agent-only setting.}
\label{append_table:relevant_analysis_agent_only_setting}
\end{table*}

\begin{table*}[]
\centering
\resizebox{\textwidth}{!}{%
\begin{tabular}{@{}cclccccccccc@{}}
\toprule
\multirow{3}{*}{\textbf{\begin{tabular}[c]{@{}c@{}}Utterance\\ type\end{tabular}}} & \multirow{3}{*}{\textbf{\begin{tabular}[c]{@{}c@{}}Persona\\ setting\end{tabular}}} &  & \multicolumn{9}{c}{\textbf{Agent only}} \\ \cmidrule(l){4-12} 
 &  &  & \multicolumn{4}{c}{\textbf{PersonaChat}} &  & \multicolumn{4}{c}{\textbf{MBTI-S2Conv}} \\ \cmidrule(lr){4-7} \cmidrule(l){9-12} 
 &  &  & Llama3.1 & Qwen2.5 & \multicolumn{1}{l}{DeepSeek-chat} & GPT4.1 &  & Llama3.1 & Qwen2.5 & \multicolumn{1}{l}{DeepSeek-chat} & GPT4.1 \\ \cmidrule(r){1-7} \cmidrule(l){9-12} 
\multirow{3}{*}{\begin{tabular}[c]{@{}c@{}}Persona\\ irrelevant\end{tabular}} & \textbf{No} &  & 57.8 & 42.1 & 76.1 & 49.0 &  & 48.8 & 47.5 & 38.9 & 45.0 \\
 & \textbf{+Noise} &  & 63.5 & 49.0 & 68.5 & 55.7 &  & 69.4 & 56.3 & 40.7 & 58.9 \\
 & \textbf{All} &  & \textbf{75.0} & \textbf{60.5} & \textbf{83.4} & \textbf{64.1} &  & \textbf{79.8} & \textbf{57.9} & 43.7 & \textbf{73.1} \\ \cmidrule(r){1-7} \cmidrule(l){9-12} 
\multirow{4}{*}{\begin{tabular}[c]{@{}c@{}}Persona\\ relevant\end{tabular}} & \textbf{No} &  & 59.2 & 44.1 & 73.2 & 51.8 &  & 104.6 & 94.6 & 112.0 & 130.2 \\
 & \textbf{Match} &  & 66.7 & 51.8 & 66.6 & 59.9 &  & 126.3 & 102.5 & \textbf{112.2} & 147.6 \\
 & \textbf{+Noise} &  & 72.7 & 56.8 & 68.6 & 64.1 &  & 133.2 & 103.9 & 111.2 & 158.7 \\
 & \textbf{All} &  & \textbf{77.9} & \textbf{63.0} & \textbf{73.7} & \textbf{66.8} &  & \textbf{148.8} & \textbf{114.6} & 101.3 & \textbf{172.5} \\ \bottomrule
\end{tabular}%
}
\caption{Response length under Agent-only setting.}
\label{append_table:Response_length_agent_only}
\end{table*}

\begin{table*}[]
\centering
\resizebox{\textwidth}{!}{%
\begin{tabular}{@{}cclccccccccc@{}}
\toprule
\multirow{3}{*}{\textbf{\begin{tabular}[c]{@{}c@{}}Utterance\\ type\end{tabular}}} & \multirow{3}{*}{\textbf{\begin{tabular}[c]{@{}c@{}}Persona\\ setting\end{tabular}}} &  & \multicolumn{9}{c}{\textbf{Partner + Agent}} \\ \cmidrule(l){4-12} 
 &  &  & \multicolumn{4}{c}{\textbf{PersonaChat}} &  & \multicolumn{4}{c}{\textbf{MBTI-S2Conv}} \\ \cmidrule(lr){4-7} \cmidrule(l){9-12} 
 &  &  & Llama3.1 & Qwen2.5 & \multicolumn{1}{l}{DeepSeek-chat} & GPT4.1 &  & Llama3.1 & Qwen2.5 & \multicolumn{1}{l}{DeepSeek-chat} & GPT4.1 \\ \cmidrule(r){1-7} \cmidrule(l){9-12} 
\multirow{3}{*}{\begin{tabular}[c]{@{}c@{}}Persona\\ irrelevant\end{tabular}} & \textbf{No} &  & 57.2 & 41.2 & \textbf{76.3} & 48.2 &  & 44.6 & 44.8 & 35.0 & 39.3 \\
 & \textbf{+Noise} &  & 59.3 & 45.7 & 74.9 & 51.5 &  & 62.3 & 51.7 & 41.8 & 50.0 \\
 & \textbf{All} &  & \textbf{73.7} & \textbf{56.4} & 75.3 & \textbf{64.5} &  & \textbf{71.1} & \textbf{56.7} & \textbf{42.2} & \textbf{70.2} \\ \cmidrule(r){1-7} \cmidrule(l){9-12} 
\multirow{4}{*}{\begin{tabular}[c]{@{}c@{}}Persona\\ relevant\end{tabular}} & \textbf{No} &  & 59.3 & 44.3 & \textbf{73.9} & 51.7 &  & 104.7 & 94.5 & 111.8 & 130.2 \\
 & \textbf{Match} &  & 62.6 & 48.4 & 64.4 & 57.8 &  & 128.0 & 101.0 & \textbf{114.3} & 156.6 \\
 & \textbf{+Noise} &  & 65.8 & 50.4 & 69.2 & 60.8 &  & 132.3 & 106.1 & 110.3 & 164.1 \\
 & \textbf{All} &  & \textbf{74.7} & \textbf{59.6} & 69.3 & \textbf{68.0} &  & \textbf{143.6} & \textbf{116.7} & 101.3 & \textbf{184.3} \\ \bottomrule
\end{tabular}%
}
\caption{Response length under Partner + Agent setting.}
\label{append_table:Response_length_agent_persona}
\end{table*}

\begin{table*}[]
\centering
\setlength{\tabcolsep}{3pt}
\resizebox{\textwidth}{!}{%
\begin{tabular}{@{}cclccccccccccccccccccc@{}}
\toprule
\multirow{3}{*}{\textbf{\begin{tabular}[c]{@{}c@{}}Utterance\\ type\end{tabular}}} & \multirow{3}{*}{\textbf{\begin{tabular}[c]{@{}c@{}}Persona\\ setting\end{tabular}}} &  & \multicolumn{9}{c}{\textbf{PersonaChat}} &  & \multicolumn{9}{c}{\textbf{MBTI-S2Conv}} \\ \cmidrule(lr){4-12} \cmidrule(l){14-22} 
 &  &  & \multirow{2}{*}{\textbf{Overuse ($\downarrow$)}} &  & \multicolumn{3}{c}{\textbf{Consistency ($\uparrow$)}} &  & \multicolumn{3}{c}{\textbf{Ref. Similarity ($\uparrow$)}} &  & \multirow{2}{*}{\textbf{Overuse ($\downarrow$)}} &  & \multicolumn{3}{c}{\textbf{Consistency ($\uparrow$)}} &  & \multicolumn{3}{c}{\textbf{Ref. Similarity ($\uparrow$)}} \\ \cmidrule(lr){6-8} \cmidrule(lr){10-12} \cmidrule(lr){16-18} \cmidrule(l){20-22} 
 &  &  &  &  & C.Score & P-Dist & P-F1 &  & BLEU & R-L & B.Score &  &  &  &  & P-Dist & P-F1 &  & BLEU & R-L & B.Score \\ \midrule
\multirow{3}{*}{Llama3.1} & \textbf{No} &  & \textbf{0.621} &  & 0.236 & 0.340 & 0.056 &  & \textbf{0.040} & \textbf{0.087} & \textbf{0.832} &  & \textbf{0.328} &  &  & 0.241 & 0.087 &  & \textbf{0.123} & \textbf{0.139} & \textbf{0.844} \\
 & \textbf{+Noise} &  & 0.824 &  & 0.523 & 0.363 & 0.060 &  & \textbf{0.040} & 0.085 & 0.831 &  & 0.546 &  &  & 0.271 & \textbf{0.100} &  & 0.106 & 0.127 & 0.841 \\
 & \textbf{All} &  & 0.948 &  & \textbf{0.913} & \textbf{0.423} & \textbf{0.063} &  & 0.039 & 0.085 & 0.830 &  & 0.718 &  &  & \textbf{0.282} & \textbf{0.100} &  & 0.101 & 0.125 & 0.840 \\ \midrule
\multirow{3}{*}{Qwen2.5} & \textbf{No} &  & \textbf{0.539} &  & 0.072 & 0.304 & 0.051 &  & \textbf{0.047} & \textbf{0.112} & \textbf{0.838} &  & \textbf{0.397} &  &  & 0.260 & 0.096 &  & \textbf{0.140} & \textbf{0.158} & \textbf{0.854} \\
 & \textbf{+Noise} &  & 0.803 &  & 0.388 & 0.341 & 0.060 &  & 0.045 & 0.104 & 0.836 &  & 0.572 &  &  & \textbf{0.271} & \textbf{0.097} &  & 0.125 & 0.149 & 0.851 \\
 & \textbf{All} &  & 0.936 &  & \textbf{0.947} & \textbf{0.401} & \textbf{0.067} &  & 0.045 & 0.102 & 0.834 &  & 0.670 &  &  & \textbf{0.271} & 0.095 &  & 0.124 & 0.154 & 0.851 \\ \midrule
\multicolumn{1}{l}{\multirow{3}{*}{DeepSeek-chat}} & \textbf{No} &  & \textbf{0.476} & \multicolumn{1}{l}{} & 0.074 & 0.326 & 0.054 & \multicolumn{1}{l}{} & 0.034 & 0.083 & 0.826 & \multicolumn{1}{l}{} & \textbf{0.268} & \multicolumn{1}{l}{} & \multicolumn{1}{l}{} & 0.198 & 0.077 & \multicolumn{1}{l}{} & \textbf{0.155} & \textbf{0.188} & \textbf{0.853} \\
\multicolumn{1}{l}{} & \textbf{+Noise} &  & 0.599 & \multicolumn{1}{l}{} & 0.240 & 0.347 & 0.056 & \multicolumn{1}{l}{} & 0.034 & 0.084 & 0.828 & \multicolumn{1}{l}{} & 0.350 & \multicolumn{1}{l}{} & \multicolumn{1}{l}{} & 0.222 & 0.082 & \multicolumn{1}{l}{} & 0.140 & 0.167 & 0.849 \\
\multicolumn{1}{l}{} & \textbf{All} &  & 0.830 & \multicolumn{1}{l}{} & \textbf{1.066} & \textbf{0.435} & \textbf{0.068} & \multicolumn{1}{l}{} & \textbf{0.036} & \textbf{0.085} & \textbf{0.830} & \multicolumn{1}{l}{} & 0.400 & \multicolumn{1}{l}{} & \multicolumn{1}{l}{} & \textbf{0.227} & \textbf{0.088} & \multicolumn{1}{l}{} & 0.138 & 0.161 & 0.850 \\ \midrule
\multirow{3}{*}{GPT4.1} & \textbf{No} &  & \textbf{0.634} &  & 0.184 & 0.333 & 0.050 &  & \textbf{0.049} & \textbf{0.114} & \textbf{0.836} &  & \textbf{0.245} &  &  & 0.201 & 0.071 &  & \textbf{0.129} & \textbf{0.178} & \textbf{0.850} \\
 & \textbf{+Noise} &  & 0.846 &  & 0.517 & 0.369 & 0.059 &  & 0.047 & 0.107 & 0.835 &  & 0.371 &  &  & 0.242 & 0.086 &  & 0.109 & 0.150 & 0.847 \\
 & \textbf{All} &  & 0.955 &  & \textbf{1.044} & \textbf{0.454} & \textbf{0.068} &  & 0.047 & 0.102 & 0.833 &  & 0.624 &  &  & \textbf{0.274} & \textbf{0.101} &  & 0.101 & 0.145 & 0.844 \\ \midrule \midrule
\multirow{3}{*}{Average} & \textbf{No} &  & \textbf{0.567} &  & 0.142 & 0.326 & 0.053 &  & \textbf{0.043} & \textbf{0.099} & \textbf{0.833} &  & \textbf{0.310} &  &  & 0.225 & 0.083 &  & \textbf{0.137} & \textbf{0.166} & \textbf{0.850} \\
 & \textbf{+Noise} &  & 0.768 &  & 0.417 & 0.355 & 0.059 &  & 0.041 & 0.095 & 0.832 &  & 0.460 &  &  & 0.251 & 0.091 &  & 0.120 & 0.148 & 0.847 \\
 & \textbf{All} &  & 0.917 &  & \textbf{0.992} & \textbf{0.428} & \textbf{0.066} &  & 0.042 & 0.093 & 0.832 &  & 0.603 &  &  & \textbf{0.264} & \textbf{0.096} &  & 0.116 & 0.146 & 0.846 \\ \bottomrule
\end{tabular}%
}
\caption{Persona-irrelevant utterances under Partner + Agent setting.}
\label{append_table:irrelevant_analysis_partner_agent_setting}
\end{table*}

\begin{table*}[]
\centering
\setlength{\tabcolsep}{3pt}
\resizebox{\textwidth}{!}{%
\begin{tabular}{@{}cclccccccccccccccccccc@{}}
\toprule
\multirow{3}{*}{\textbf{\begin{tabular}[c]{@{}c@{}}Utterance\\ type\end{tabular}}} & \multirow{3}{*}{\textbf{\begin{tabular}[c]{@{}c@{}}Persona\\ setting\end{tabular}}} &  & \multicolumn{9}{c}{\textbf{PersonaChat}} &  & \multicolumn{9}{c}{\textbf{MBTI-S2Conv}} \\ \cmidrule(lr){4-12} \cmidrule(l){14-22} 
 &  &  & \multirow{2}{*}{\textbf{Overuse ($\downarrow$)}} &  & \multicolumn{3}{c}{\textbf{Consistency ($\uparrow$)}} &  & \multicolumn{3}{c}{\textbf{Ref. Similarity ($\uparrow$)}} &  & \multirow{2}{*}{\textbf{Overuse ($\downarrow$)}} &  & \multicolumn{3}{c}{\textbf{Consistency ($\uparrow$)}} &  & \multicolumn{3}{c}{\textbf{Ref. Similarity ($\uparrow$)}} \\ \cmidrule(lr){6-8} \cmidrule(lr){10-12} \cmidrule(lr){16-18} \cmidrule(l){20-22} 
 &  &  &  &  & C.Score & P-Dist & P-F1 &  & BLEU & R-L & B.Score &  &  &  &  & P-Dist & P-F1 &  & BLEU & R-L & B.Score \\ \midrule
\multirow{4}{*}{Llama3.1} & \textbf{No} &  & \textbf{0.736} &  & 0.421 & 0.371 & 0.057 &  & 0.041 & 0.086 & \textbf{0.833} &  & \textbf{0.861} &  &  & 0.344 & \textbf{0.124} &  & \textbf{0.114} & \textbf{0.122} & \textbf{0.848} \\
 & \textbf{Match} &  & 0.754 &  & 0.793 & 0.393 & 0.060 &  & \textbf{0.043} & \textbf{0.089} & \textbf{0.833} &  & 0.875 &  &  & 0.358 & 0.122 &  & 0.105 & 0.118 & 0.845 \\
 & \textbf{+Noise} &  & 0.807 &  & 0.839 & 0.403 & 0.062 &  & \textbf{0.043} & 0.087 & 0.832 &  & 0.889 &  &  & 0.362 & 0.122 &  & 0.103 & 0.117 & 0.845 \\
 & \textbf{All} &  & 0.890 &  & \textbf{0.978} & \textbf{0.437} & \textbf{0.065} &  & 0.041 & 0.084 & 0.831 &  & 0.918 &  &  & \textbf{0.370} & 0.122 &  & 0.099 & 0.115 & 0.845 \\ \midrule
\multirow{4}{*}{Qwen2.5} & \textbf{No} &  & \textbf{0.656} &  & 0.308 & 0.342 & 0.058 &  & 0.046 & 0.097 & \textbf{0.837} &  & \textbf{0.858} &  &  & 0.336 & \textbf{0.115} &  & \textbf{0.113} & \textbf{0.122} & \textbf{0.856} \\
 & \textbf{Match} &  & 0.747 &  & 0.721 & 0.376 & 0.066 &  & 0.047 & 0.099 & \textbf{0.837} &  & 0.868 &  &  & 0.340 & 0.114 &  & 0.109 & 0.121 & 0.855 \\
 & \textbf{+Noise} &  & 0.794 &  & 0.810 & 0.386 & 0.068 &  & \textbf{0.048} & \textbf{0.100} & 0.836 &  & 0.884 &  &  & 0.344 & 0.114 &  & 0.106 & 0.118 & 0.854 \\
 & \textbf{All} &  & 0.881 &  & \textbf{1.079} & \textbf{0.425} & \textbf{0.069} &  & 0.046 & 0.096 & 0.835 &  & 0.897 &  &  & \textbf{0.349} & 0.112 &  & 0.101 & 0.116 & 0.853 \\ \midrule
\multirow{4}{*}{DeepSeek-chat} & \textbf{No} &  & \textbf{0.578} &  & 0.416 & 0.366 & 0.051 &  & 0.037 & 0.080 & 0.830 &  & \textbf{0.825} &  &  & 0.343 & 0.120 &  & 0.113 & 0.132 & 0.859 \\
 & \textbf{Match} &  & 0.674 &  & 0.805 & 0.388 & 0.061 &  & \textbf{0.041} & \textbf{0.090} & \textbf{0.833} &  & 0.838 &  &  & \textbf{0.349} & 0.123 &  & 0.111 & 0.129 & 0.857 \\
 & \textbf{+Noise} &  & 0.743 &  & 0.923 & 0.410 & 0.062 &  & 0.039 & 0.084 & 0.832 &  & 0.849 &  &  & 0.348 & \textbf{0.124} &  & 0.113 & 0.130 & 0.858 \\
 & \textbf{All} &  & 0.795 &  & \textbf{1.081} & \textbf{0.439} & \textbf{0.067} &  & \textbf{0.041} & \textbf{0.090} & \textbf{0.833} &  & 0.847 &  &  & 0.342 & \textbf{0.124} &  & \textbf{0.120} & \textbf{0.134} & \textbf{0.860} \\ \midrule
\multirow{4}{*}{GPT4.1} & \textbf{No} &  & \textbf{0.731} & \multicolumn{1}{l}{} & 0.467 & 0.372 & 0.057 & \multicolumn{1}{l}{} & \textbf{0.050} & \textbf{0.098} & \textbf{0.835} & \multicolumn{1}{l}{} & \textbf{0.889} & \multicolumn{1}{l}{} & \multicolumn{1}{l}{} & 0.361 & \textbf{0.116} & \multicolumn{1}{l}{} & \textbf{0.094} & \textbf{0.111} & \textbf{0.851} \\
 & \textbf{Match} &  & 0.812 & \multicolumn{1}{l}{} & 1.052 & 0.417 & 0.066 & \multicolumn{1}{l}{} & \textbf{0.050} & \textbf{0.098} & 0.834 & \multicolumn{1}{l}{} & 0.898 & \multicolumn{1}{l}{} & \multicolumn{1}{l}{} & 0.376 & 0.113 & \multicolumn{1}{l}{} & 0.086 & 0.102 & 0.845 \\
 & \textbf{+Noise} &  & 0.858 & \multicolumn{1}{l}{} & 1.114 & 0.433 & 0.069 & \multicolumn{1}{l}{} & 0.047 & 0.097 & 0.834 & \multicolumn{1}{l}{} & 0.909 & \multicolumn{1}{l}{} & \multicolumn{1}{l}{} & 0.380 & 0.112 & \multicolumn{1}{l}{} & 0.082 & 0.100 & 0.844 \\
 & \textbf{All} &  & 0.920 & \multicolumn{1}{l}{} & \textbf{1.226} & \textbf{0.470} & \textbf{0.071} & \multicolumn{1}{l}{} & 0.047 & 0.094 & 0.833 & \multicolumn{1}{l}{} & 0.931 & \multicolumn{1}{l}{} & \multicolumn{1}{l}{} & \textbf{0.390} & 0.110 & \multicolumn{1}{l}{} & 0.077 & 0.096 & 0.841 \\ \midrule \midrule
\multirow{4}{*}{Average} & \textbf{No} &  & \textbf{0.675} &  & 0.403 & 0.363 & 0.056 &  & 0.044 & 0.090 & \textbf{0.834} &  & \textbf{0.858} &  &  & 0.346 & \textbf{0.119} &  & \textbf{0.108} & \textbf{0.122} & \textbf{0.854} \\
 & \textbf{Match} &  & 0.747 &  & 0.843 & 0.394 & 0.063 &  & \textbf{0.045} & \textbf{0.094} & \textbf{0.834} &  & 0.870 &  &  & 0.356 & 0.118 &  & 0.103 & 0.118 & 0.851 \\
 & \textbf{+Noise} &  & 0.800 &  & 0.922 & 0.408 & 0.065 &  & 0.044 & 0.092 & \textbf{0.834} &  & 0.883 &  &  & 0.358 & 0.118 &  & 0.101 & 0.116 & 0.850 \\
 & \textbf{All} &  & 0.871 &  & \textbf{1.091} & \textbf{0.443} & \textbf{0.068} &  & 0.044 & 0.091 & 0.833 &  & 0.898 &  &  & \textbf{0.363} & 0.117 &  & 0.099 & 0.115 & 0.850 \\ \bottomrule
\end{tabular}%
}
\caption{Persona-relevant utterances under Partner + Agent setting.}
\label{append_table:relevant_analysis_partner_agent_setting}
\end{table*}

\begin{table*}[]
\centering
\resizebox{0.9\textwidth}{!}{%
\begin{tabular}{@{}ccccccccc@{}}
\toprule
\multirow{2}{*}{Model} & \multirow{2}{*}{Method} & \multicolumn{3}{c}{PersonaChat} &  & \multicolumn{3}{c}{MBTI-S2Conv} \\ \cmidrule(lr){3-5} \cmidrule(l){7-9} 
 &  & Overuse ($\downarrow$) & ROUGE-L ($\uparrow$) & PAS ($\uparrow$) &  & Overuse ($\downarrow$) & ROUGE-L ($\uparrow$) & PAS ($\uparrow$) \\ \midrule
\multirow{6}{*}{Llama3.1} & Vanilla & 0.917 & 0.086 & 0.049 &  & 0.861 & 0.120 & 0.322 \\
 & CoT & 0.867 & 0.090 & 0.128 &  & 0.819 & 0.120 & 0.402 \\
 & Decomp. & 0.899 & 0.090 & 0.081 &  & 0.864 & 0.116 & 0.326 \\
 & Self-refine & 0.913 & 0.079 & 0.086 &  & 0.848 & 0.111 & 0.335 \\
 & CAA & {\ul 0.830} & {\ul 0.091} & {\ul 0.224} &  & {\ul 0.816} & {\ul 0.127} & {\ul 0.423} \\
 & SCONPOS & \textbf{0.743} & \textbf{0.103} & \textbf{0.301} &  & \textbf{0.722} & \textbf{0.135} & \textbf{0.540} \\ \midrule
\multirow{6}{*}{Qwen2.5} & Vanilla & 0.880 & 0.103 & 0.138 &  & 0.827 & 0.129 & 0.354 \\
 & CoT & 0.880 & 0.092 & 0.137 &  & {\ul 0.772} & 0.127 & 0.429 \\
 & Decomp. & 0.883 & 0.091 & 0.104 &  & 0.776 & 0.115 & 0.434 \\
 & Self-refine & 0.889 & {\ul 0.093} & 0.122 &  & 0.825 & {\ul 0.137} & 0.390 \\
 & CAA & {\ul 0.774} & 0.092 & {\ul 0.317} &  & 0.788 & \textbf{0.139} & 0.435 \\
 & SCONPOS & \textbf{0.718} & \textbf{0.111} & \textbf{0.345} &  & \textbf{0.746} & 0.136 & \textbf{0.455} \\ \midrule
\multirow{6}{*}{DeepSeek-chat} & Vanilla & 0.777 & 0.098 & 0.285 &  & 0.728 & {\ul 0.156} & 0.480 \\
 & CoT & {\ul 0.730} & 0.093 & {\ul 0.338} &  & 0.733 & 0.150 & 0.490 \\
 & Decomp. & 0.746 & 0.099 & 0.331 &  & {\ul 0.712} & {\ul 0.156} & {\ul 0.510} \\
 & Self-refine & 0.775 & 0.098 & 0.326 &  & 0.760 & 0.155 & 0.455 \\
 & CAA & 0.734 & \textbf{0.107} & 0.317 &  & 0.741 & \textbf{0.157} & 0.470 \\
 & SCONPOS & \textbf{0.496} & {\ul 0.105} & \textbf{0.486} &  & \textbf{0.656} & 0.154 & \textbf{0.528} \\ \bottomrule
\end{tabular}%
}
\caption{Performance comparison results across datasets and models.}
\label{tab:full_performance_comparison}
\end{table*}

\begin{table*}[]
\resizebox{\textwidth}{!}{%
\begin{tabular}{@{}cclllllll@{}}
\toprule
\multirow{2}{*}{Model} & \multirow{2}{*}{Method} & \multicolumn{3}{c}{PersonaChat} & \multicolumn{1}{c}{} & \multicolumn{3}{c}{MBTI-S2Conv} \\ \cmidrule(l){3-9} 
 &  & \multicolumn{1}{c}{Overuse ($\downarrow$)} & \multicolumn{1}{c}{ROUGE-L ($\uparrow$)} & \multicolumn{1}{c}{PAS ($\uparrow$)} & \multicolumn{1}{c}{} & \multicolumn{1}{c}{Overuse ($\downarrow$)} & \multicolumn{1}{c}{ROUGE-L ($\uparrow$)} & \multicolumn{1}{c}{PAS ($\uparrow$)} \\ \midrule
\multirow{4}{*}{Llama3.1} & Vanilla & 0.92$\to$0.74 (-19\%) & 0.09$\to$0.10 (+19\%) & 0.05$\to$0.30 (+515\%) &  & 0.86$\to$0.72 (-16\%) & 0.12$\to$0.14 (+12\%) & 0.32$\to$0.54 (+68\%) \\
 & CoT & 0.87$\to$0.70 (-20\%) & 0.09$\to$0.09 (+5\%) & 0.13$\to$0.32 (+149\%) &  & 0.82$\to$0.73 (-11\%) & 0.12$\to$0.14 (+14\%) & 0.40$\to$0.52 (+29\%) \\
 & Decomp. & 0.90$\to$0.62 (-31\%) & 0.09$\to$0.09 (+5\%) & 0.08$\to$0.42 (+418\%) &  & 0.86$\to$0.74 (-14\%) & 0.12$\to$0.14 (+20\%) & 0.33$\to$0.48 (+47\%) \\
 & Self-refine & 0.91$\to$0.68 (-26\%) & 0.08$\to$0.08 (+6\%) & 0.09$\to$0.35 (+305\%) &  & 0.85$\to$0.73 (-14\%) & 0.11$\to$0.13 (+15\%) & 0.34$\to$0.53 (+58\%) \\ \midrule
\multirow{4}{*}{Qwen2.5} & Vanilla & 0.88$\to$0.72 (-19\%) & 0.10$\to$0.11 (+7\%) & 0.14$\to$0.34 (+150\%) &  & 0.83$\to$0.75 (-10\%) & 0.13$\to$0.14 (+5\%) & 0.35$\to$0.45 (+28\%) \\
 & CoT & 0.88$\to$0.71 (-19\%) & 0.09$\to$0.11 (+15\%) & 0.14$\to$0.34 (+144\%) &  & 0.77$\to$0.72 (-6\%) & 0.13$\to$0.14 (+7\%) & 0.43$\to$0.46 (+8\%) \\
 & Decomp. & 0.88$\to$0.65 (-27\%) & 0.09$\to$0.10 (+11\%) & 0.10$\to$0.40 (+282\%) &  & 0.78$\to$0.70 (-9\%) & 0.12$\to$0.14 (+26\%) & 0.43$\to$0.50 (+16\%) \\
 & Self-refine & 0.89$\to$0.73 (-17\%) & 0.09$\to$0.11 (+13\%) & 0.12$\to$0.34 (+182\%) &  & 0.82$\to$0.73 (-11\%) & 0.14$\to$0.15 (+9\%) & 0.39$\to$0.47 (+21\%) \\ \midrule
\multirow{4}{*}{DeepSeek-chat} & Vanilla & 0.78$\to$0.50 (-36\%) & 0.10$\to$0.11 (+8\%) & 0.28$\to$0.49 (+71\%) &  & 0.73$\to$0.66 (-10\%) & 0.16$\to$0.15 (-1\%) & 0.48$\to$0.53 (+10\%) \\
 & CoT & 0.73$\to$0.48 (-34\%) & 0.09$\to$0.09 (-1\%) & 0.34$\to$0.53 (+57\%) &  & 0.73$\to$0.67 (-8\%) & 0.15$\to$0.15 (+1\%) & 0.49$\to$0.52 (+7\%) \\
 & Decomp. & 0.75$\to$0.46 (-38\%) & 0.10$\to$0.10 (+3\%) & 0.33$\to$0.53 (+61\%) &  & 0.71$\to$0.64 (-11\%) & 0.16$\to$0.15 (-4\%) & 0.51$\to$0.57 (+11\%) \\
 & Self-refine & 0.78$\to$0.50 (-36\%) & 0.10$\to$0.10 (+3\%) & 0.33$\to$0.52 (+61\%) &  & 0.76$\to$0.67 (-12\%) & 0.15$\to$0.16 (+5\%) & 0.46$\to$0.54 (+19\%) \\ \bottomrule
\end{tabular}%
}
\caption{Impact of SCONPOS. Each cell is formatted as ``w/o steering $\to$ w/ steering (relative change (\%))''. The steering vector is extracted from only the Vanilla setting and applied to all methods.}
\label{tab:CoT_steering_effect}
\end{table*}

\begin{table*}[]
\centering
\resizebox{0.9\textwidth}{!}{%
\begin{tabular}{llllllll}
\hline
\multicolumn{1}{c}{\multirow{2}{*}{Model}} & \multicolumn{3}{c}{MBTI-S2Conv $\to$ PersonaChat} & \multicolumn{1}{c}{} & \multicolumn{3}{c}{PersonaChat $\to$ MBTI-S2Conv} \\ \cline{2-8} 
\multicolumn{1}{c}{} & Overuse ($\downarrow$) & ROUGE-L ($\uparrow$) & PAS ($\uparrow$) &  & Overuse ($\downarrow$) & ROUGE-L ($\uparrow$) & PAS ($\uparrow$) \\ \hline
Llama3.1 & 0.917 & 0.086 & 0.049 &  & 0.861 & 0.120 & 0.322 \\
$\quad$ +SCONPOS & 0.811 (-12\%) & 0.111 (+28\%) & 0.206 (+320\%) &  & 0.791 (-8\%) & 0.135 (+12\%) & 0.455 (+41\%) \\
$\quad$ + CAA & 0.901 (-2\%) & 0.094 (+9\%) & 0.049 (-0\%) &  & 0.825 (-4\%) & 0.117 (-3\%) & 0.385 (+19\%) \\ \hline
Qwen2.5 & 0.880 & 0.103 & 0.138 &  & 0.827 & 0.129 & 0.354 \\
$\quad$ +SCONPOS & 0.769 (-12\%) & 0.122 (+20\%) & 0.266 (+93\%) &  & 0.797 (-4\%) & 0.143 (+11\%) & 0.379 (+9\%) \\
$\quad$ + CAA & 0.886 (+1\%) & 0.103 (-1\%) & 0.159 (+14\%) &  & 0.749 (-9\%) & 0.129 (-0\%) & 0.437 (+23\%) \\ \hline
DeepSeek-chat & 0.777 & 0.098 & 0.285 &  & 0.728 & 0.156 & 0.480 \\
$\quad$ +SCONPOS & 0.668 (-14\%) & 0.110 (+13\%) & 0.386 (+36\%) &  & 0.646 (-11\%) & 0.152 (-2\%) & 0.513 (+7\%) \\
$\quad$ + CAA & 0.762 (-2\%) & 0.098 (+0\%) & 0.304 (+7\%) &  & 0.684 (-6\%) & 0.161 (+3\%) & 0.475 (-1\%) \\ \hline
\end{tabular}%
}
\caption{Transferability of SCONPOS steering vectors across datasets. The steering vector is extracted from the Vanilla setting of one dataset and applied to the other.}
\label{tab:cross_dataset}
\end{table*}

\begin{figure*}
    \centering
    \includegraphics[width=\textwidth]{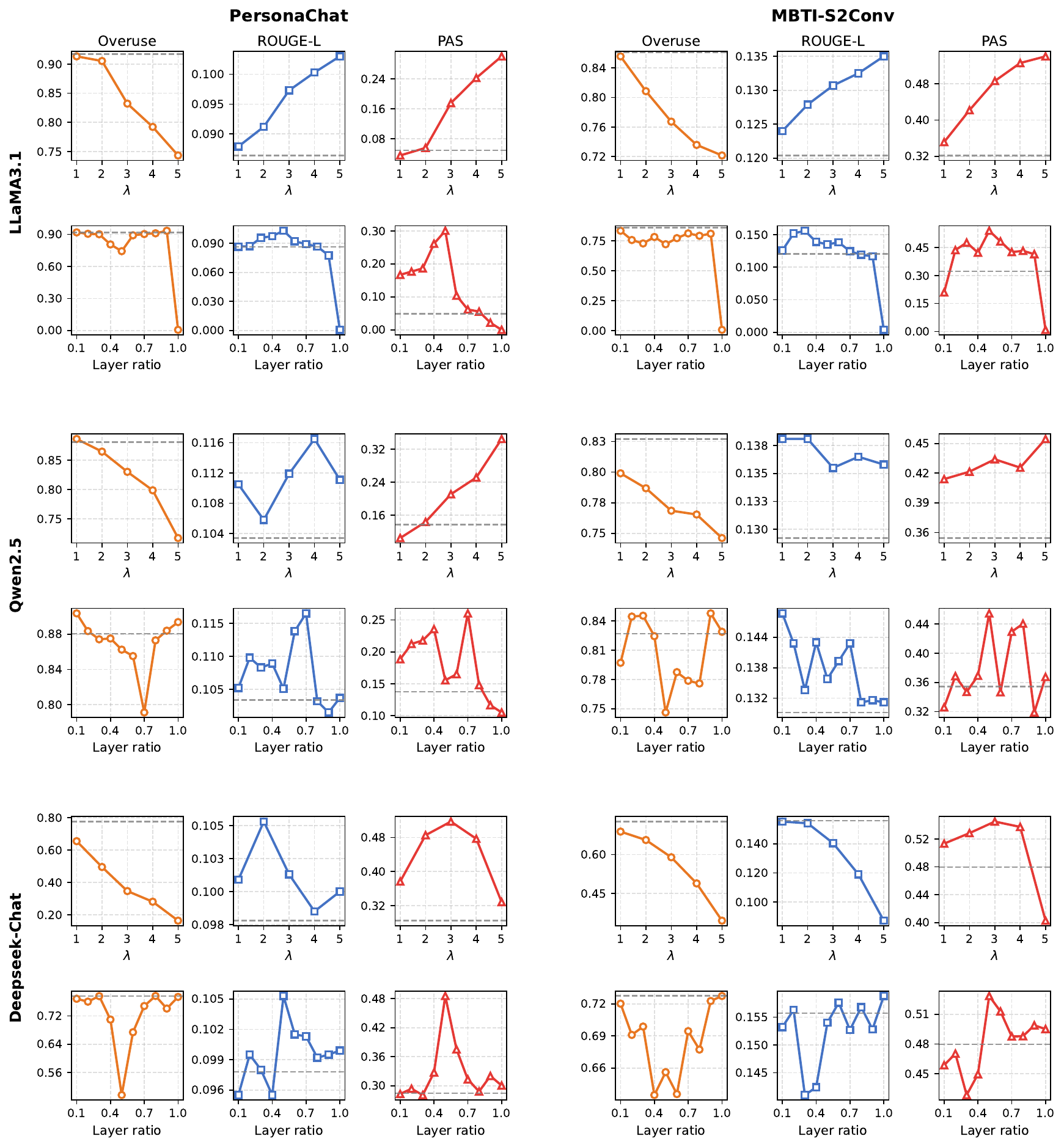}
    \caption{Performance results across suppression intensity $\lambda$ and target layer ratio for LLaMA3.1, Qwen2.5, and Deepseek-chat on PersonaChat and MBTI-S2Conv.}
    \label{fig:ablation_combined}
\end{figure*}

\begin{table*}[h]
\centering
\small
\resizebox{\textwidth}{!}{%
\begin{tabular}{lllc}
\toprule
\textbf{Type} & \textbf{Artifact} & \textbf{License} & \textbf{Link} \\
\midrule
\multirow{2}{*}{Dataset}
 & PersonaChat & Unknown & \url{https://github.com/facebookresearch/ParlAI/tree/main/projects/personachat} \\
 & MBTI-S2Conv & Unknown & \url{https://github.com/morecry/CharacterChat} \\
 \midrule
\multirow{3}{*}{Model}
 & Llama-3.1-8B-Instruct & Apache-2.0 & \url{https://huggingface.co/meta-llama/Llama-3.1-8B-Instruct} \\
 & Qwen2.5-7B-Instruct & Apache-2.0 & \url{https://huggingface.co/Qwen/Qwen2.5-7B-Instruct} \\
 & DeepSeek-llm-7b-chat & MIT & \url{https://huggingface.co/deepseek-ai/deepseek-llm-7b-chat} \\
\midrule
\multirow{5}{*}{Software}
 & HuggingFace Transformers & Apache-2.0 & \url{https://github.com/huggingface/transformers} \\
 & Scikit-learn & BSD-3-Clause & \url{https://scikit-learn.org/stable/} \\
 & Scipy & BSD-3-Clause & \url{https://docs.scipy.org/doc/scipy/index.html} \\
 & Matplotlib & PSF & \url{https://matplotlib.org/stable/project/license.html} \\
 & PyTorch & \href{https://github.com/pytorch/pytorch?tab=License-1-ov-file\#readme}{License link} & \url{https://github.com/pytorch/pytorch} \\
\bottomrule
\end{tabular}
}
\caption{Artifacts used in this work and their licenses and links.}
\label{tab:artifacts}
\end{table*}

\begin{figure*}[h]
    \begin{tcolorbox}[title={Vanilla}]

    \textbf{\textless System Prompt\textgreater} \\
    - Your persona:\\
    \{List of persona attributes\} \\
    - Dialogue history:\\
    \{List of dialogue history\} \\
    - - - \\
    You aim to have a dialogue with the partner, maintaining your persona.\\
    Respond to the partner's utterance considering the given dialogue history between your partner and you.\\
    Also, when responding to the partner's utterance, you can refer to the given persona attributes.\\
        
    \textbf{\textless User Prompt\textgreater} \\
    Partner: \{Last utterance\}
    \end{tcolorbox}
    \caption{Vanilla prompt for PersonaChat.}
    \label{fig:prompt_vanilla}

\end{figure*}

\begin{figure*}[h]
    \begin{tcolorbox}[title={Vanilla}]

    \textbf{\textless System Prompt\textgreater} \\
    - Your persona:\\
    \{List of persona attributes\} \\
    - Dialogue history:\\
    \{List of dialogue history\} \\
    - - - \\
    You aim to chat with partner. The partner has encountered some problems recently, and you will give comfort and help to the partner through chatting.\\
    Respond to the partner's utterance considering the given dialogue history between your partner and you.\\
    Do not reveal in any way that you are an AI or that you are roleplaying, always remember that you are who you are.\\
        
    \textbf{\textless User Prompt\textgreater} \\
    Partner: \{Last utterance\}
    \end{tcolorbox}
    \caption{Vanilla prompt for MBTI-S2Conv.}
    \label{fig:prompt_vanilla_mbti}

\end{figure*}

\begin{figure*}[h]
    \begin{tcolorbox}[title={Chain-of-Thought (CoT)}]

    \textbf{\textless System Prompt\textgreater} \\
    - Your persona:\\
    \{List of persona attributes\} \\
    - Dialogue history:\\
    \{List of dialogue history\} \\
    - - - \\
    You aim to have a dialogue with the partner, maintaining your persona.\\
    Respond to the partner's utterance considering the given dialogue history between your partner and you.\\
    Also, when responding to the partner's utterance, you can refer to the given persona attributes.\\

    First, you should generate reasoning path for an appropriate response. Let's think step-by-step.\\
    
    - Reasoning path:\\
    \{generated reasoning path\} \\
    
    Respond appropriately referring to the given reasoning path.\\
        
    \textbf{\textless User Prompt\textgreater} \\
    Partner: \{Last utterance\}
    \end{tcolorbox}
    \caption{Prompt for Chain-of-Thought.}
    \label{fig:prompt_cot}

\end{figure*}

\begin{figure*}[h]
    \begin{tcolorbox}[title={Task Decomposition}]

    \textbf{\textless System Prompt\textgreater} \\
    - Your persona:\\
    \{List of persona attributes\} \\
    - Dialogue history:\\
    \{List of dialogue history\} \\
    - - - \\
    You aim to have a dialogue with the partner, maintaining your persona.\\
    Respond to the partner's utterance considering the given dialogue history between your partner and you.\\
    Also, when responding to the partner's utterance, you can refer to the given persona attributes.\\

    First, you should decompose the given claim into sub-claims for an appropriate response. Let's break down the claim!\\
    
    - Sub-claims:\\
    \{generated sub-claims\} \\
    
    Respond appropriately referring to the given sub-claims.\\
        
    \textbf{\textless User Prompt\textgreater} \\
    Partner: \{Last utterance\}
    \end{tcolorbox}
    \caption{Prompt for Task Decomposition.}
    \label{fig:prompt_td}

\end{figure*}

\begin{figure*}[h]
    \begin{tcolorbox}[title={Evaluation Rubric for Persona Appropriateness}]

    - Agent persona:\\
    \{List of persona attributes\} \\
    - Dialogue history:\\
    \{List of dialogue history\} \\
    - Last utterance:\\
    \{Partner's last utterance\} \\
    - Response:\\
    \{Agent's response\} \\
    - - - \\
    You aim to evaluate whether the persona attributes are appropriately used in the response.\\
    You should assign one score from 1 to 5, based on the following criteria:\\

    Score=1: The response only uses persona attributes that are inappropriate for the dialogue last utterance and dialogue history.\\
    Score=2: The response fails to use persona attributes that are appropriate for the last utterance and dialogue history, even though such persona attribute exists.\\
    Score = 3: The response mixes appropriate and inappropriate persona attributes with respect to the last utterance and dialogue history.\\
    Score = 4: The response does not use any persona attributes, when all persona attributes are inappropriate for the last utterance and dialogue history.\\
    Score = 5: The response only uses persona attributes that are appropriate for the dialogue last utterance and dialogue history.
            
    \end{tcolorbox}
    \caption{Evaluation rubric of persona appropriateness for evaluating metric.}
    \label{fig:rubric_pa_PAS}

\end{figure*}

\begin{figure*}[h]
    \begin{tcolorbox}[title={Pairwise Evaluation Rubric for Persona Appropriateness}]

    - Agent persona:\\
    \{List of persona attributes\} \\
    - Dialogue history:\\
    \{List of dialogue history\} \\
    - Last utterance:\\
    \{Partner's last utterance\} \\
    - Response A:\\
    \{Agent's response A\} \\
    - Response B:\\
    \{Agent's response B\} \\
    - - - \\
    You aim to compare the two responses in terms of persona appropriateness.
    A response is persona-appropriate if it uses persona attributes that are relevant to the last utterance and dialogue history, while avoiding persona attributes that are irrelevant to the dialogue context.\\

    Choose one of the following labels:\\

    A: Response A uses persona attributes more appropriately than Response B.\\
    B: Response B uses persona attributes more appropriately than Response A.\\
            
    \end{tcolorbox}
    \caption{Pairwise evaluation rubric of persona appropriateness for method comparison.}
    \label{fig:rubric_pa_method}

\end{figure*}

\begin{figure*}[h]
    \begin{tcolorbox}[title={Evaluation Rubric for Dialogue Coherence}]

    - Agent persona:\\
    \{List of persona attributes\} \\
    - Dialogue history:\\
    \{List of dialogue history\} \\
    - Last utterance:\\
    \{Partner's last utterance\} \\
    - Response:\\
    \{Agent's response\} \\
    - - - \\
    You aim to evaluate whether the response is coherence with dialogue history and natural with last utterance.\\
    You should assign one score from 1 to 5, based on the following criteria:\\

    Score = 1: The response is completely incoherence with dialogue history and unnatural with last utterance.\\
    Score = 2: The response is slightly incoherence with dialogue history and unnatural with last utterance.\\
    Score = 3: The response is moderately coherence with dialogue history and natural with last utterance.\\
    Score = 4: The response is highly coherence with dialogue history and natural with last utterance.\\
    Score = 5: The response is perfectly coherence with dialogue history and natural with last utterance.
            
    \end{tcolorbox}
    \caption{Evaluation rubric of dialogue coherence for evaluating metric.}
    \label{fig:rubric_dc_PAS}

\end{figure*}

\begin{figure*}[h]
    \begin{tcolorbox}[title={Pairwise Evaluation Rubric for Dialogue Coherence}]

    - Agent persona:\\
    \{List of persona attributes\} \\
    - Dialogue history:\\
    \{List of dialogue history\} \\
    - Last utterance:\\
    \{Partner's last utterance\} \\
    - Response A:\\
    \{Agent's response A\} \\
    - Response B:\\
    \{Agent's response B\} \\
    - - - \\
    Your task is to determine which response is more coherent with the dialogue history and more natural as a reply to the last utterance. 
    A response is more dialogue-coherent if it follows the previous conversation flow, responds appropriately to the partner's last utterance, and sounds natural in the given dialogue context.\\

    Choose one of the following labels:\\

    A: Response A is more coherent and natural than Response B.\\
    B: Response B is more coherent and natural than Response A.\\
            
    \end{tcolorbox}
    \caption{Pairwise evaluation rubric of dialogue coherence for method comparison.}
    \label{fig:rubric_dc_method}

\end{figure*}

\clearpage

\end{document}